%% file: main-arxiv.tex
\PassOptionsToPackage{dvipsnames}{xcolor}
\documentclass[11pt]{article}

\usepackage[margin=1in]{geometry}
\usepackage{times}
\usepackage{natbib}
\input{math_commands.tex}

\input{package.tex}

\title{ChartRevise: A Dataset and Evaluation Protocol for Exact Chart Editing via Code}
\author{Jiaxiang Tang\\
  University of Minnesota, Twin-Cities\\
  \texttt{tang0836@umn.edu}
  \and
  Yi Zhou\\
  IBM Research\\
  \texttt{yi.zhou@ibm.com}
  \and
  Chad DeLuca\\
 IBM Research\\
  \texttt{delucac@us.ibm.com}
  \and
  Rogerio Feris\\
  IBM Research\\
  \texttt{rsferis@us.ibm.com}
  \and
  Ahmed Khalil Omrani\\
  Horizon School of Digital Technologies\\
  \texttt{ahmed.khalil.omrani@horizon-tech.tn}\\
  \and
  Zhi-Li Zhang\\
  University of Minnesota, Twin-Cities\\
  \texttt{zhzhang@cs.umn.edu}
  \and
  Pengyuan Li\thanks{Co-corresponding author.}\\
  IBM Research\\
  \texttt{pengyuan@ibm.com}
  \and
  Ali Anwar\footnotemark[1]\\
  University of Minnesota, Twin-Cities\\
  \texttt{aanwar@umn.edu}
}

\date{}

\begin{document}
\maketitle

\begin{abstract}
Chart editing requires cross-modal edit grounding, realizing a requested visual
change in the code that draws it, with necessary related updates and without
altering unrelated content. Existing benchmarks emphasize either code
executability or chart quality, but their metrics do not clearly distinguish
request completion from missed coupled updates and gratuitous changes. We
introduce ChartRevise, a structured dataset and evaluation protocol for exact
program-grounded chart editing. For dataset construction, we build on the
grammar of graphics to systematically cover chart-editing operations, using
source-program checks to verify their applicability across chart types and
libraries. To improve edit exactness, our pipeline checks individual
requirements and guides repair or exclusion when they are unmet. The resulting
dataset contains 92{,}438 records covering 344 edit types across 20 chart types
and three plotting libraries. For evaluation, our reference-free protocol
separately measures atomic requirement completion, identifies gratuitous
changes, and detects missed coupled updates. These checks are combined with
successful execution and rendering to determine exact-edit success. Across
five models and four external benchmarks, fine-tuning yields relative gains of
16\% in mean requirement recall and 22\% in mean exact-edit rate.
\end{abstract}

\input{1-intro}

\input{2-related}
\input{3-taxonomy}

\input{5-evaluation}

\input{6-conclusion}

\bibliographystyle{plainnat}
\bibliography{iclr2027_conference}

\clearpage
\appendix
\input{99-appendix}

\end{document}

%% file: math_commands.tex
\usepackage{amsmath,amsfonts,bm}

\def\eqref#1{equation~\ref{#1}}

\def\1{\bm{1}}

\DeclareMathAlphabet{\mathsfit}{\encodingdefault}{\sfdefault}{m}{sl}
\SetMathAlphabet{\mathsfit}{bold}{\encodingdefault}{\sfdefault}{bx}{n}



%% file: package.tex
\newcommand{\sysname}{ChartRevise\xspace}

\newcommand{\NCells}{4{,}316}

\newcommand{\NChartTypes}{20}
\newcommand{\NCategories}{twelve}

\newcommand{\NReleased}{90{,}589}
\newcommand{\NReleasedTrain}{\NReleased}
\newcommand{\NReleasedTest}{1{,}849}
\newcommand{\NReleasedTotal}{92{,}438}

\newcommand{\NReleasedCompound}{4{,}519}

\newcommand{\NEditsAll}{344}              

\newcommand{\vtt}[1]{\texttt{\small #1}}

\usepackage{xspace}
\usepackage[dvipsnames]{xcolor}

\newcommand{\gain}[1]{\textcolor{ForestGreen}{#1}}
\newcommand{\loss}[1]{\textcolor{BrickRed}{#1}}
\newcommand{\tie}[1]{\textcolor{gray}{#1}}
\newcommand{\dd}[2]{\shortstack{#1\\[-1pt]{\scriptsize #2}}}
\usepackage{listings}
\usepackage{pifont}
\newcommand{\cmark}{\ding{51}} 
\usepackage{hyperref}
\usepackage{url}
\usepackage{graphicx}
\usepackage{booktabs}
\usepackage{array,tabularx,colortbl}
\usepackage{multirow}
\usepackage{placeins}
\usepackage{cleveref}
\usepackage[most]{tcolorbox}

%% file: 1-intro.tex
\section{Introduction}
\label{sec:intro}


Charts are often created from plotting programs, so editing a chart usually requires editing its code. The user, however, asks for a change to the rendered chart rather than to the code, so the requested change is defined on the chart that the program draws. Chart editing therefore requires cross-modal edit grounding, connecting a text instruction to the visual components it refers to and to the code regions that draw them. We study this program-grounded setting, in which a model receives a rendered chart, its source program, and an instruction, and must modify the right code regions so that the rendered chart reflects the requested change while unrelated chart properties remain unchanged. Errors occur in two directions. Under-editing leaves an \emph{explicit requirement} unmet or misses a \emph{coupled update} that the requested change implies, and over-editing introduces \emph{gratuitous changes} to unrelated properties. Coupled updates are never stated in the instruction and gratuitous changes lie outside it, so a check that asks only whether the instruction was followed catches neither.

\begin{figure}[!ht]
  \centering
  \includegraphics[width=\linewidth]{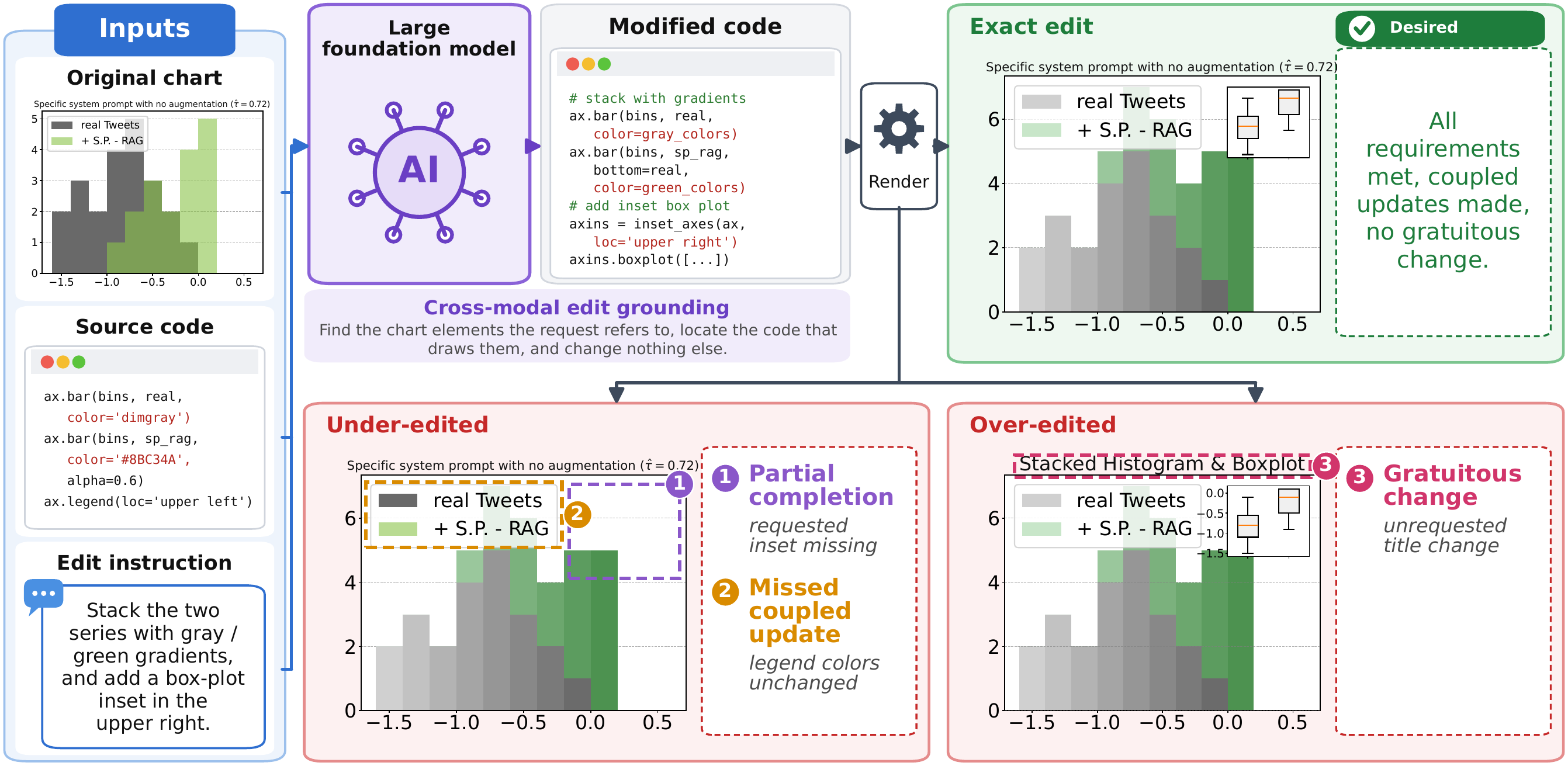}
    \vspace{-.3cm}
  \caption{\footnotesize Failure modes in Chart2Code-L2 task 0565. Under-edited and exact panels are constructed; the over-edited panel is the released reference.}
  \label{fig:teaser}
  \vspace{-.3cm}
\end{figure}

\Cref{fig:teaser} illustrates both directions on a Chart2Code-L2 task~\citep{tang2026charts} requesting stacked bars with gray/green gradients and an inset boxplot in the upper right. The under-edited panel stacks and recolors the bars but omits the boxplot, leaving an explicit requirement unmet in the code (\emph{partial completion}), and keeps the old legend colors although the legend depends on the recolored bars (\emph{missed coupled update}). The over-edited panel completes the request, but also rewrites the title, a component the instruction never refers to (\emph{gratuitous change}). Because references serve as both training targets and evaluation standards, such a reference can teach and reward over-editing, so exactness must be enforced per requirement in both training data and evaluation.

Prior work builds instruction-driven editing systems \citep{yan2024chartreformer,goswami2025plotedit} and iterative code generation for scientific plots \citep{yang2024matplotagent}, and benchmarks cover validated edits~\citep{zhao2025chartedit}, multimodal instructions~\citep{yang2025chartm3}, and complex modifications~\citep{kapadnis2026charteditbench,tang2026charts}. However, two gaps remain. 
First, existing resources provide limited coverage of editing operations across chart types and plotting libraries. FigEdit\citep{li2026charts} fixes a set of operations over chart specifications and ChartM$^3$ builds its pool around Matplotlib APIs \citep{yang2025chartm3}, so coverage reflects curation and does not extend systematically to new chart objects or libraries. Second, evaluation does not distinguish errors. Aggregate quality scores hide which requirements were met and what else changed, and reference-based scores can inherit flaws in the reference. Benchmarks check instruction following \citep{kapadnis2026charteditbench}, preservation\citep{yang2025chartm3,li2026charts}, or geometric consistency \citep{yu2026chartsync} cover parts of the task, but none jointly scores completion, coupled updates, and gratuitous changes.

We introduce \sysname{}, a dataset and evaluation protocol for exact program-grounded chart editing, built on the idea that each edit is a set of atomic requirements that can be independently checked. For dataset construction, we derive object--action templates from the grammar of graphics and instantiate them only when their source conditions hold, yielding single and compound requests across three plotting libraries. Execution, completion, and preservation checks with atomic audits filter or repair each record. For evaluation, a reference-free protocol decomposes each edit into requirements and verifies them directly in the chart-generating program, without a ground-truth program. It separately scores \emph{completion}, \emph{consistency} of coupled updates, and \emph{preservation} of unrelated content; an edit is exact only if it executes and passes all three checks. Fine-tuning on \sysname{} consistently improves completion and coupled updates on external benchmarks, but less often reduces gratuitous changes, suggesting that supervised edits teach models what to edit more reliably than what to preserve.
%
Our contributions are summarized as follows:
\begin{itemize}\itemsep0pt \parskip0pt \topsep0pt
  \item \textbf{Grammar-derived editing dataset.} Edit types are derived from the grammar of graphics, which makes coverage systematic and extensible to new objects, chart types, and libraries. The release contains \NReleasedTotal{} records spanning \NEditsAll{} edit types in \NCategories{} categories, \NChartTypes{} chart types, three plotting libraries, and compound requests of up to twelve actions (\Cref{sec:grammar,sec:change-boundary,sec:release}).
  \item \textbf{Fine-grained construction audits.} After initial validation for completion and preservation, a judge from a different model family audits
coupled updates in every record, and compound records are further audited requirement by requirement rather than by one overall judge. Failed checks guide targeted repair or exclusion, and a human audit finds 84\% accepted-set precision (\Cref{sec:auditing,sec:release}).
  \item \textbf{Reference-free exact-edit evaluation.} To our knowledge, this is the first protocol that jointly scores explicit requirements, coupled updates against a fixed per-task inventory, and gratuitous changes, without a reference edit. A human audit finds 93\% agreement with the judge (\Cref{sec:exact-edit-protocol,sec:human-transfer}).
  \item \textbf{Transfer and editing-behavior analysis.} Across five models and four external benchmarks, fine-tuning on \sysname{} improves requirement recall and coupled-update recall in 19 of 20 comparisons and exact-edit rate in 17, but reduces gratuitous changes in only 10 (\Cref{sec:task-outcomes}).
\end{itemize}

%% file: 2-related.tex
\section{Related Works}
\label{sec:related}
\vspace{-0.1cm}
We compare \sysname{} with existing works, and highlight \sysname{}'s combination of a large-scale editing dataset,
atomic auditing and repair, and joint checks of requirement completion, preservation,
and coupled updates in \Cref{tab:related}.

\begin{table*}[t]
\vspace{-0.6cm}
\centering
\caption{\footnotesize Task scope, construction checks, and explicit evaluation checks of chart-editing resources.}
\label{tab:related}
\begingroup
\fontsize{8}{10}\selectfont
\setlength{\tabcolsep}{2pt}
\renewcommand{\arraystretch}{1.16}
\renewcommand{\tabularxcolumn}[1]{m{#1}}
\newcommand{\relatedref}[1]{{\fontsize{7}{8}\selectfont\color{black!60}#1}}
\arrayrulecolor{black!55}
\begin{tabularx}{\linewidth}{@{}l r c
  >{\hsize=.80\hsize\linewidth=\hsize\centering\arraybackslash}X
  >{\hsize=1.70\hsize\linewidth=\hsize\centering\arraybackslash}X
  >{\hsize=.95\hsize\linewidth=\hsize\centering\arraybackslash}X
  >{\hsize=.90\hsize\linewidth=\hsize\centering\arraybackslash}X
  >{\hsize=.65\hsize\linewidth=\hsize\centering\arraybackslash}X@{}}
\toprule[0.7pt]
\multirow{2}{*}{\textbf{Resource}}
  & \multicolumn{3}{c}{\textbf{Dataset scope}}
  & \multirow{2}{*}{\shortstack{\textbf{Construction}\\\textbf{checks}}}
  & \multicolumn{3}{c}{\textbf{Evaluation checks}} \\
\cmidrule(lr){2-4}\cmidrule(l){6-8}
  & \textbf{Tasks} & \shortstack{\textbf{Chart}\\\textbf{types}}
  & \shortstack{\textbf{Instruction}\\\textbf{types}}
  & & \shortstack{\textbf{Atomic}\\\textbf{requirements}}
  & \textbf{Preservation}
  & \shortstack{\textbf{Coupled}\\\textbf{updates}} \\
\midrule
ChartEdit
  & 1,405 & 19 & 6
  & \shortstack{Human\\(request, code, chart)} & --- & --- & --- \\
\addlinespace[3pt]
ChartM$^3$
  & \shortstack[r]{1K eval.\\24K train} & 10 & ---
  & Model (charts) & --- & \shortstack{\cmark\\\relatedref{(code-space)}} & --- \\
\addlinespace[3pt]
ChartEditVista
  & 7,964 & 31 & 6
  & \shortstack{Model (charts)\\Human (code, charts)} & --- & --- & --- \\
\addlinespace[3pt]
ChartEditBench
  & 4,142 & 37 & 35
  & \shortstack{Model (code, charts)\\Program (assertions)} & \shortstack{\cmark\\\relatedref{(assertions)}} & --- & --- \\
\addlinespace[3pt]
Chart2Code L2
  & 1,010 & 19 & ---
  & --- & --- & --- & --- \\
\addlinespace[3pt]
FigEdit
  & 30,836 & 10 & 9
  & Program (specs) & --- & \shortstack{\cmark\\\relatedref{(image-space)}} & --- \\
\addlinespace[3pt]
ChartSync
  & 870 & 9 & ---
  & \shortstack{Model (code, charts)\\Human (code, charts)} & --- & \shortstack{\cmark\\\relatedref{(image-space)}} & \shortstack{\cmark\\\relatedref{(geometry)}} \\
\midrule
\rowcolor{blue!5}
\textbf{\sysname{}}
  & \textbf{\NReleasedTotal{}} & \textbf{\NChartTypes{}} & \textbf{\NEditsAll{}}
  & \shortstack{\textbf{Model(code,charts)}\\[2pt]
    \textbf{Atomic audit+repair}\\[2pt]
    \textbf{Human audit}}
  & \cmark & \shortstack{\cmark\\\relatedref{(code audit)}} & \cmark \\
\bottomrule[0.7pt]
\end{tabularx}
\endgroup
\vspace{-.25cm}
\end{table*}

\textbf{Chart-editing resources.}
Existing datasets study program revision and multi-turn editing
\citep{zhao2025chartedit,tang2026charts,kapadnis2026charteditbench}, multimodal
instructions \citep{yang2025chartm3,chen2026charteditor}, and direct image editing
\citep{li2026charts,yu2026chartsync}. 
ChartReformer \citep{yan2024chartreformer} edits charts from images and natural-language requests, while PlotEdit \citep{goswami2025plotedit} applies
multimodal agents to chart editing in PDFs.
These resources cover diverse interaction formats, but their instructions typically draw on predefined edits to chart objects or library-specific APIs.
Such inventories offer limited support for systematic extension across plotting
libraries, whereas our grammar-derived taxonomy provides shared object-action
templates that can be extended with new objects and actions, and adapted across charts and libraries.

\textbf{Evaluation.}
Current benchmarks report code or chart quality scores\citep{zhao2025chartedit} and, in some cases,
separate instruction-following and preservation scores
\citep{tang2026charts,yang2025chartm3,
li2026charts,yu2026chartsync,ku2024viescore,ku2024imagenhub}.
These aggregate measures can hide individual omissions and unrequested changes.
ChartEditBench \citep{kapadnis2026charteditbench} checks individual conditions through assertions for programmatic
instructions, but passing these assertions does not establish that all necessary
related updates are complete.
ChartSync explicitly checks text-to-geometry dependencies, but these checks cover
a specific class of coupled updates rather than their broader range \citep{yu2026chartsync}.
Thus, satisfying those reported criteria does not necessarily establish that every explicit requirement and necessary related update is complete without gratuitous
changes. An execution failure also leaves unresolved how much of the request was
completed. Following work on evaluating LLM judges, we separately assess agreement
with human judgments \citep{zheng2023judging,liu2023g}. These limitations leave cross-modal edit grounding only partially
assessed: whether the requested visual changes are implemented on the
intended chart components, with necessary dependent updates and no
unrelated changes. Our protocol checks these outcomes in the source
and edited programs at the requirement level separately,
including in programs that fail to execute. \Cref{app:extended-related} provides extended discussions on related works.

%% file: 3-taxonomy.tex
\vspace{-0.1cm}
\section{Systematic Construction and Validation of \sysname{}}
\label{sec:taxonomy}
\vspace{-0.1cm}
We construct \sysname{} to support cross-modal edit grounding through systematic
operation coverage and verified editing examples. The construction combines a grammar-derived taxonomy,
source-program applicability checks, and multi-stage validation.

\textbf{Cross-modal edit grounding.} For each editing example, the model must connect the requested visual change to
the relevant chart component and its implementation in code, then realize the requested
state and necessary dependent updates while preserving unrelated properties.
An \emph{explicit requirement} specifies a visual property requested by the instruction;
a \emph{coupled update} is an unstated change needed to keep dependent chart elements
consistent. A \emph{gratuitous change} modifies a chart property outside the scope of
the explicit requirements and necessary coupled updates. An edit is \emph{exact} when the program executes
and renders, satisfies all explicit requirements and coupled updates, and introduces no
gratuitous change. \Cref{sec:exact-edit-protocol} defines how these criteria are measured.

\begin{figure*}[!t]
\vspace{-0.25cm}
  \centering
  \includegraphics[width=.75\textwidth]{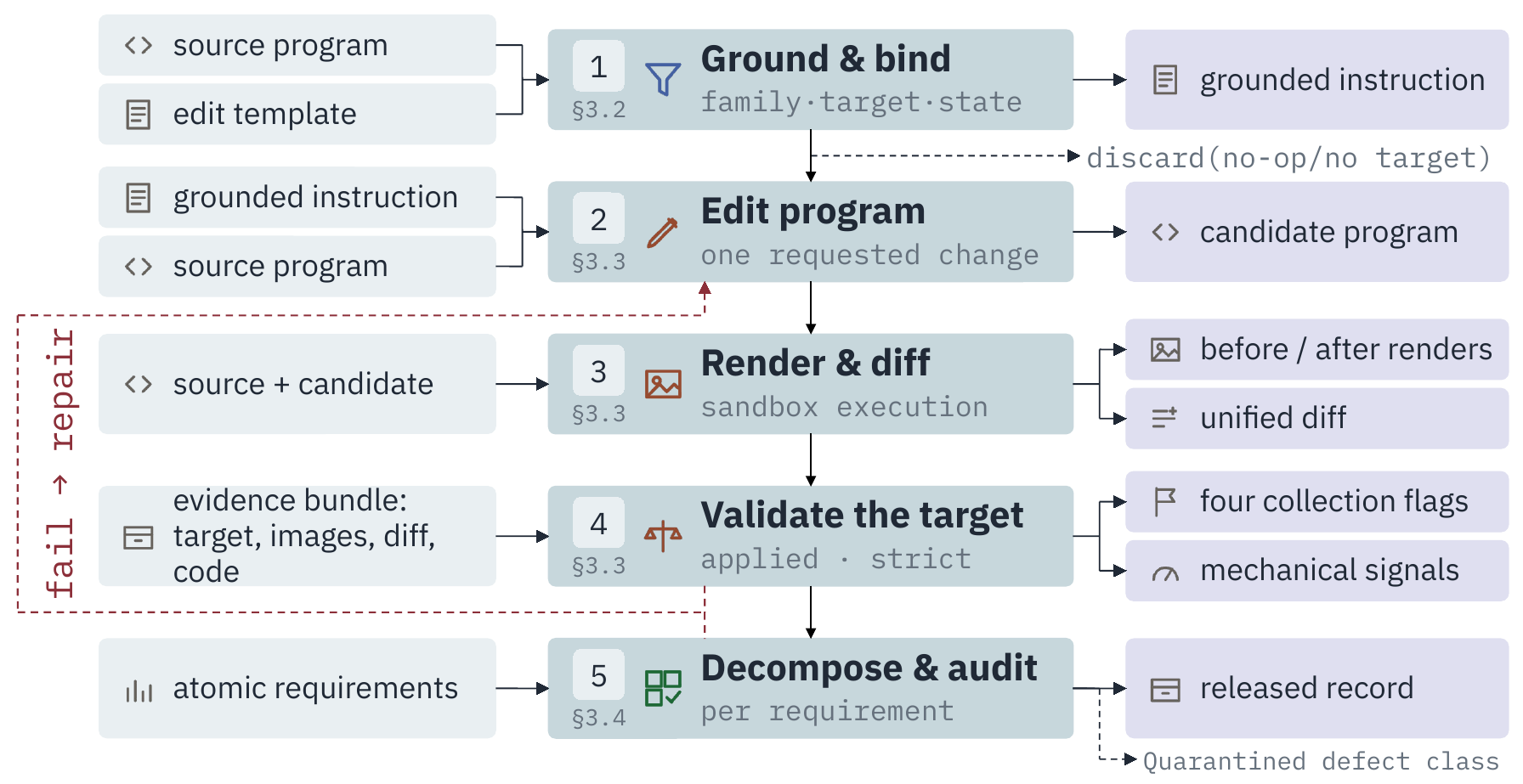}
  \vspace{-0.5cm}
  \caption{Dataset construction pipeline. Source-grounded instructions guide program
  generation. Execution and target validation produce repair feedback, and post-processing audits assess cross-modal edit grounding by checking
completion, consistency, and preservation before release.}
  \label{fig:pipeline}
  \vspace{-.3cm}
\end{figure*}

Our pipeline turns edit types from the taxonomy into verified training examples
using existing chart programs.  \Cref{fig:pipeline} summarizes its five steps.
Step~1 selects an applicable template from the taxonomy and binds its targets and
arguments to a source program, producing a source-grounded instruction.
Step~2 rewrites the program under this instruction.
Step~3 executes and renders the source and candidate programs.
Step~4 validates the target edit using code and visual evidence; execution and validation
failures provide feedback for repair.
Step~5 decomposes the instruction into atomic requirements and audits completion and
coupled updates before release.


\subsection{Grammar-derived edit taxonomy}
\label{sec:grammar}
To define the dataset's edit scope, we select \NCategories{} operation categories
based on the grammar of graphics, its layered formulation, and view composition
\citep{wilkinson2011grammar,wickham2010layered,satyanarayan2016vega,satyanarayan2015reactive}.
These categories cover data and transforms, encodings, scales and guides,
marks and layers, position, coordinates, facets, annotations, and style.

Existing resources construct edit inventories from selected object edits or
library-specific operation pools
\citep{zhao2025chartedit,yang2025chartm3,li2026charts}.
To support \emph{systematic coverage}, we pair editable objects with applicable
actions within each grammar-derived category, forming templates with placeholders. For example, legend--rename yields ``Change the legend title
to \texttt{\{title\}}.'' These semantic templates can be reused across compatible
chart types and plotting libraries.
\Cref{app:grammar,app:grounding} provide the object inventory and instantiated examples.

\subsection{Applicability, source grounding, and composition}
\label{sec:change-boundary}
An edit template is useful only when the source program supports the requested
operation. We therefore check compatibility, bind valid targets, and
combine instantiated actions into requests.

We first construct a compatibility table whose rows are edit templates and whose columns
are chart types. A cell marks a template as applicable to a chart type when charts of that type can contain the template's target object, giving \NCells{} compatible combinations (\Cref{app:coverage}). 
For each source program, we first use this table to filter templates by chart
type, then inspect the code to verify that the required target object or plotting context is present. For instance, a legend-move template requires identifying an existing
legend, whereas adding a reference line requires an axes object to host it.
The final collection covers \NEditsAll{} applicable edit types.

Next, we instantiate each applicable template against the source program. Existing field names,
series labels, and target objects are bound to the template, while new values such as a
position or font size are selected under its constraints. We reject no-op requests and
inconsistent bindings. We call the resulting instruction \emph{source-grounded}, as
its target or plotting context and arguments are resolved for that source program.
\Cref{app:grounding} provides more examples.

A single-action request instantiates one template. To construct compound requests, we
combine instantiated actions for the same source chart, using 4--6 actions in tier~2 and
8--12 in tier~3, and check for direct conflicts among them. This produces requests that
require several changes to be completed together. \Cref{app:compound} provides additional details on compound request construction.

\subsection{Edited program generation, validation, and repair}
\label{sec:construction}

Given a source-grounded instruction, the successful edit must realize the requested chart
states without disturbing unrelated properties. Therefore, after generating candidate programs, we validate these outcomes and repair failures using diagnostic feedback.

{\textbf{Generation.}}
\label{sec:grounding}
We draw source programs for different chart types from ChartNet~\citep{kondic2026chartnet}. We use \texttt{gemma-4-31B-it}\citep{gemmateam2026gemma4} to obtain a complete edited program based on the source chart, source program and the source-grounded instruction. Its prompt specifies the requested change and requires unrelated chart content to be preserved. 

{\textbf{Validation.}}
We execute and render the source and edited programs in the same sandbox. A judge model, \texttt{gemma-4-31B-it} , then checks each candidate against the before/after renders, the unified code diff, and the complete edited program to verify that code changes affect the intended visual component and preserve unrelated chart properties. A candidate is marked applied when the requested edit reaches the named target, and strict when it also executes and makes no unrequested change, movement, or removal.

{\textbf{Repair.}}
Validation failures provide the feedback for repair. Execution or rendering failures return a
traceback to the editor, while an unapplied edit returns the judge's rationale. Repaired candidates are executed, rendered, and validated again before acceptance. Candidates that apply the requested edit but fail preservation are retained as diagnostic records for further processing. Detailed rules and implementation for generation and validation are described in \Cref{app:construction}.

\subsection{Post-processing audits}
\label{sec:auditing}
An overall judgment can overlook missing operations and misclassify necessary coupled
updates. A second judge from a different model family, \texttt{Qwen3.8-27B} \citep{qwen38}, therefore
re-audits the records before release and decides the final acceptance.

{\textbf{Coupled updates.}} A requested change often forces dependent elements to change, such as the legend entry of a removed series. The audit checks these dependent elements in every record, single-action or compound. A record that leaves a dependent element in its old state is repaired with this feedback and excluded if the repair fails. Conversely, a necessary dependent change is not counted as a gratuitous change, so records rejected for such changes in first pass are restored.

{\textbf{Compound requirements.}} Among compound records with an initial judgment, 34.6\% had been marked applied although at least one requested operation was not carried out. The audit decomposes each compound instruction into atomic requirements and checks each using the same validation criteria as single-action edits.
Consequently, consistency and preservation are assessed across the union of atomic requirements.
Unmet requirements guide targeted repair, and records that still fail are excluded. Details of post-processing audits are given in \Cref{app:compound}.

\begin{figure}[!t]
  \centering
  \includegraphics[width=0.8\linewidth]{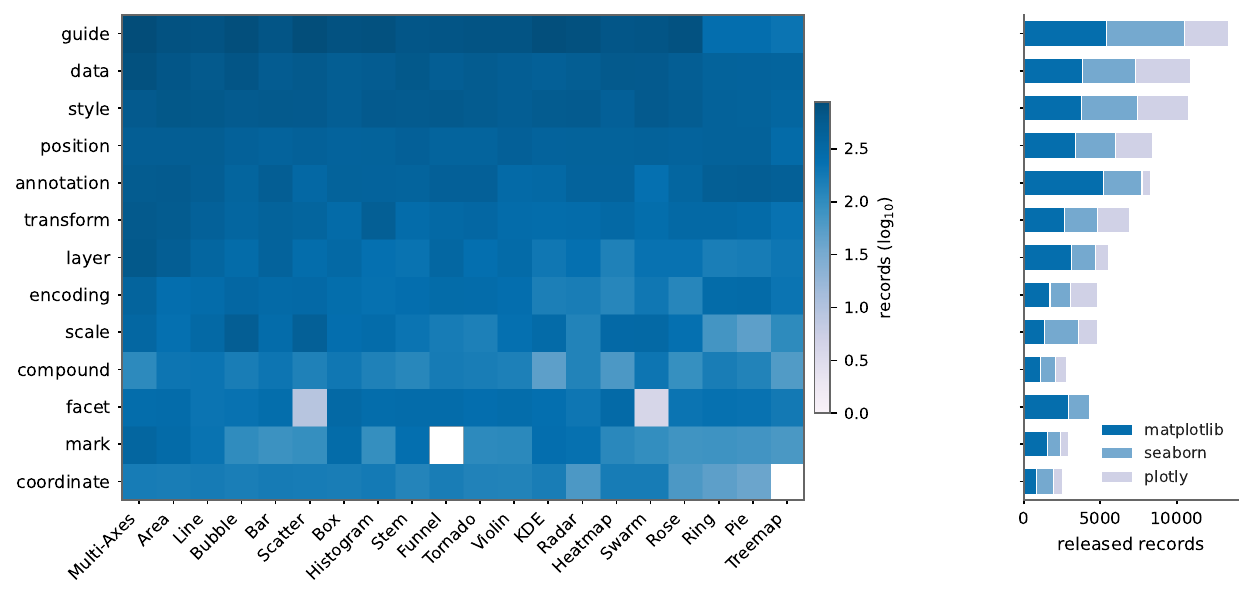}
  \vspace{-.3cm}
  \caption{\footnotesize Record counts aggregated by category and chart type (left) and plotting library (right).}
  \label{fig:coverage}
  \vspace{-.3cm}
\end{figure}

\begin{figure}
    \centering
    \includegraphics[width=\linewidth]{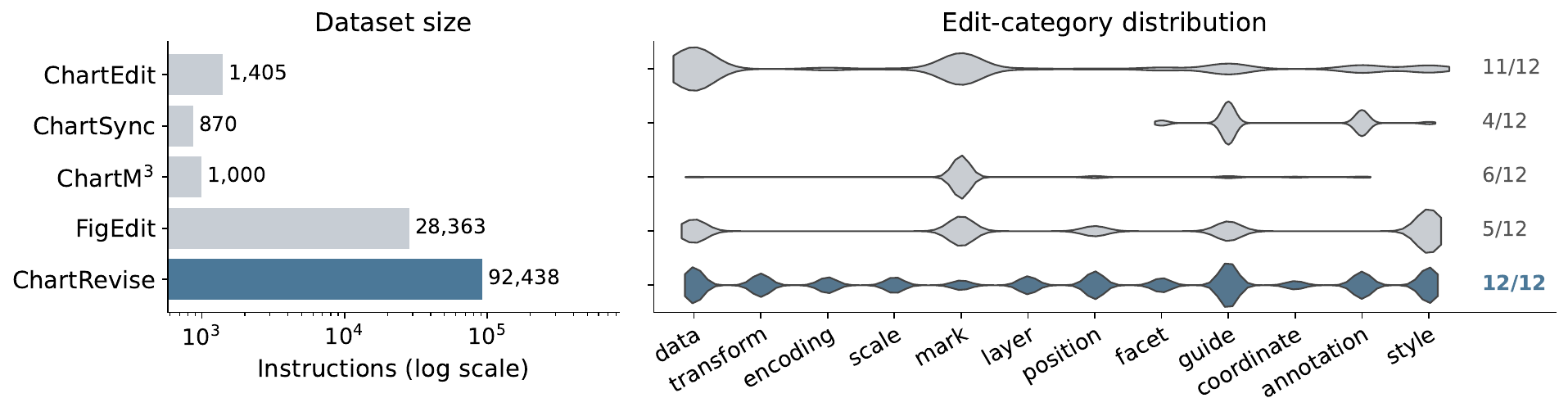}
    \vspace{-.4cm}
    \caption{Dataset size and edit-category coverage under our taxonomy.}
    \label{fig:coverage-comparison}
    \vspace{-.2cm}
\end{figure}

\subsection{Released dataset and quality assessment}\label{sec:release}

We characterize its collection outcomes, then use human audit to
estimate the correctness of accepted records.

\sysname{} contains \NReleasedTrain{} training and \NReleasedTest{} test records.
\Cref{fig:coverage} shows that editing examples span diverse chart types and
all three plotting libraries, rather than concentrating on a few chart settings.
Under the category mapping in \Cref{fig:coverage-comparison}, \sysname{} contains more
instructions than the displayed resources and spans all \NCategories{} categories more consistently, while the other
resources span four to eleven. Detailed collection outcomes appear in \Cref{app:dataset-quality}.
Human audit of 500 accepted training records estimates release-weighted
task-level correctness at 84.2\%. Requirement-level correctness reaches 95.6\%
over the 1{,}071 requirements in the 200 audited compound tasks.

%% file: 5-evaluation.tex
\section{Evaluation}
\label{sec:finetune}
Fine-tuning on \sysname{} improves requirement completion, coupled-update consistency,
and exact-edit success on external benchmarks, while preservation gains are less consistent.
We describe the setup and evaluation protocol, report transfer results, analyze the
remaining failures, and assess agreement with human judgments.

\textbf{Setup.}\label{sec:eval-setup}
We evaluate Phi-3.5-Vision-Instruct{~\citep{phi3}}, InternVL2.5-8B{~\citep{chen2024internvl}}, Qwen3.5-4B and 9B~\citep{qwen3.5}, and
Granite-Vision-4.1-4B~\citep{granite-vision-4.1-4b} on ChartEdit{~\citep{zhao2025chartedit}}, Chart2Code Level~2 (editing level only){~\citep{tang2026charts}}, ChartM$^3${~\citep{yang2025chartm3}}, and ChartSync{~\citep{yu2026chartsync}}.
None of these benchmarks contributes source charts, instructions, or programs to training.
To measure external transfer, we compare each base checkpoint with its fine-tuned counterpart
on the same tasks. Models receive the rendered source chart, the complete source program, and the
instruction, and generate a complete edited program. Every output is
executed in a common rendering sandbox. Judge-based quantities use
\texttt{gemma-4-31B-it}~\citep{gemmateam2026gemma4} at temperature zero. Training details, benchmark adaptations, and complete
benchmark-facing results appear in \Cref{app:config,app:ood-benchmark-facing}.

\subsection{Reference-free evaluation protocol}
\label{sec:exact-edit-protocol}

We measure completion, preservation, and consistency using the shared edit criteria
in \Cref{sec:taxonomy}, without a ground-truth edited program. Each task has fixed lists of
explicit requirements and necessary coupled updates, shared across models. A judge checks
each required chart state in the source and edited programs, including whether the edit
reaches the intended element and uses valid APIs.

The per-task \emph{completion ratio} (CR) is the fraction of explicit requirements
satisfied. \emph{Requirement recall} (CR$_\mu$) is the satisfied fraction pooled over all requirements;
\emph{full-completion rate} is the fraction of tasks satisfying every explicit requirement.
\emph{Coupled-update recall} is the fraction of required dependent states satisfied, pooled over inventory items.
\emph{Gratuitous changes} are neither requested nor necessary coupled updates and are
reported as a mean count per task.

For task $i$, let $s_i$ of $K_i$ explicit requirements be satisfied, $g_i$ count gratuitous
changes, and $c_i$ count missed coupled updates. Let $e_i=1$ if the edited program executes
and renders successfully, and $e_i=0$ otherwise. Thus, completion ratio of task $i$ is $\mathrm{CR}_i=\frac{s_i}{K_i}$, and requirement recall is
$\mathrm{CR}_{\mu}=\frac{\sum_i s_i}{\sum_i K_i}$. Exact editing requires all four conditions,
\[
\mathrm{Exact}_i
=\mathbf{1}[e_i=1]\,\mathbf{1}[s_i=K_i]\,\mathbf{1}[g_i=0]\,\mathbf{1}[c_i=0].
\]
Exact-edit rate is averaged over all tasks. Requirement satisfaction, gratuitous changes, and missed coupled updates are made separately
from execution, so correctly implemented changes can still be measured in programs
that fail to run. Prompts and implementation details are given in \Cref{app:exact-prompts}.

\subsection{Main transfer results}
\label{sec:task-outcomes}

\textbf{Completion and coupled updates improve more consistently than preservation.}
Fine-tuning improves requirement recall and coupled-update recall in 19 of 20
model--benchmark comparisons and exact-edit rate in 17, but reduces gratuitous changes
in only 10 (\Cref{tab:main}).

\begin{table*}[!ht]
\vspace{-.5cm}
\centering
\scriptsize
\setlength{\tabcolsep}{3pt}
\renewcommand{\arraystretch}{0.9}
\caption{External transfer under the {evaluation} protocol. \gain{Green} numbers mark improvements over the base model, and \loss{red} numbers mark declines. $rel.\ \Delta$ measures the mean change across models.}
\label{tab:main}
\resizebox{\linewidth}{!}{
\begin{tabular}{llrrrrr}
\toprule
Benchmark & Family & \shortstack{Requirement\\recall (CR$_\mu$)} $\uparrow$ & \shortstack{Full\\completion} $\uparrow$ & \shortstack{Gratuitous\\per task} $\downarrow$ & \shortstack{Coupled-update\\recall} $\uparrow$ & \shortstack{Exact-edit\\rate} $\uparrow$ \\
\midrule
\multirow{6}{*}{\shortstack[l]{\textbf{ChartEdit}\\{\tiny 1{,}405 tasks}}}
 & Phi-3.5-V & \textbf{.773}\,\gain{+.138} & \textbf{.674}\,\gain{+.155} & \textbf{.346}\,\gain{-.205} & \textbf{.540}\,\gain{+.076} & \textbf{.425}\,\gain{+.119} \\
 & InternVL2.5 & \textbf{.707}\,\gain{+.234} & \textbf{.593}\,\gain{+.215} & \textbf{.373}\,\gain{-.007} & \textbf{.500}\,\gain{+.188} & \textbf{.378}\,\gain{+.162} \\
 & Qwen3.5-4B & \textbf{.867}\,\gain{+.047} & \textbf{.804}\,\gain{+.050} & \textbf{.165}\,\gain{-.091} & \textbf{.688}\,\gain{+.004} & \textbf{.572}\,\gain{+.055} \\
 & Qwen3.5-9B & \textbf{.895}\,\gain{+.068} & \textbf{.846}\,\gain{+.085} & .135\,\loss{+.021} & \textbf{.731}\,\gain{+.064} & \textbf{.616}\,\gain{+.032} \\
 & Granite & \textbf{.812}\,\gain{+.114} & \textbf{.733}\,\gain{+.128} & \textbf{.459}\,\gain{-.482} & \textbf{.622}\,\gain{+.077} & \textbf{.439}\,\gain{+.168} \\
 & \textit{rel.\ $\Delta$ of mean} & \textit{\textcolor{ForestGreen}{+17.4\%}} & \textit{\textcolor{ForestGreen}{+21.0\%}} & \textit{\textcolor{ForestGreen}{-34.1\%}} & \textit{\textcolor{ForestGreen}{+15.3\%}} & \textit{\textcolor{ForestGreen}{+28.3\%}} \\
\cmidrule(lr){1-7}
\multirow{6}{*}{\shortstack[l]{\textbf{Chart2Code-L2}\\{\tiny 1{,}010 tasks}}}
 & Phi-3.5-V & \textbf{.691}\,\gain{+.120} & \textbf{.032}\,\gain{+.020} & 1.433\,\loss{+.150} & \textbf{.493}\,\gain{+.182} & \textbf{.002}\,\gain{+.001} \\
 & InternVL2.5 & \textbf{.624}\,\gain{+.199} & \textbf{.022}\,\gain{+.019} & 1.185\,\loss{+.400} & \textbf{.440}\,\gain{+.272} & \textbf{.001}\,\gain{+.001} \\
 & Qwen3.5-4B & \textbf{.844}\,\gain{+.013} & \textbf{.171}\,\gain{+.039} & \textbf{.900}\,\gain{-.051} & \textbf{.714}\,\gain{+.041} & \textbf{.044}\,\gain{+.009} \\
 & Qwen3.5-9B & \textbf{.896}\,\gain{+.053} & \textbf{.277}\,\gain{+.077} & .749\,\loss{+.043} & \textbf{.772}\,\gain{+.124} & \textbf{.090}\,\gain{+.017} \\
 & Granite & \textbf{.723}\,\gain{+.206} & \textbf{.040}\,\gain{+.028} & 1.314\,\loss{+.196} & \textbf{.563}\,\gain{+.278} & .000\,\loss{-.002} \\
 & \textit{rel.\ $\Delta$ of mean} & \textit{\textcolor{ForestGreen}{+18.5\%}} & \textit{\textcolor{ForestGreen}{+51.0\%}} & \textit{\textcolor{BrickRed}{+15.2\%}} & \textit{\textcolor{ForestGreen}{+43.0\%}} & \textit{\textcolor{ForestGreen}{+23.4\%}} \\
\cmidrule(lr){1-7}
\multirow{6}{*}{\shortstack[l]{\textbf{ChartM$^3$}\\{\tiny 1{,}000 tasks}}}
 & Phi-3.5-V & \textbf{.390}\,\gain{+.097} & \textbf{.321}\,\gain{+.121} & \textbf{.926}\,\gain{-.370} & \textbf{.348}\,\gain{+.094} & \textbf{.152}\,\gain{+.087} \\
 & InternVL2.5 & \textbf{.338}\,\gain{+.088} & \textbf{.246}\,\gain{+.073} & .809\,\loss{+.162} & \textbf{.388}\,\gain{+.153} & \textbf{.154}\,\gain{+.069} \\
 & Qwen3.5-4B & \textbf{.623}\,\gain{+.150} & \textbf{.540}\,\gain{+.125} & \textbf{.407}\,\gain{-.112} & \textbf{.553}\,\gain{+.045} & \textbf{.331}\,\gain{+.121} \\
 & Qwen3.5-9B & \textbf{.701}\,\gain{+.181} & \textbf{.607}\,\gain{+.145} & \textbf{.268}\,\gain{-.021} & \textbf{.628}\,\gain{+.251} & \textbf{.378}\,\gain{+.089} \\
 & Granite & \textbf{.409}\,\gain{+.041} & \textbf{.347}\,\gain{+.069} & \textbf{.741}\,\gain{-.445} & \textbf{.439}\,\gain{+.030} & \textbf{.188}\,\gain{+.082} \\
 & \textit{rel.\ $\Delta$ of mean} & \textit{\textcolor{ForestGreen}{+29.3\%}} & \textit{\textcolor{ForestGreen}{+34.9\%}} & \textit{\textcolor{ForestGreen}{-20.0\%}} & \textit{\textcolor{ForestGreen}{+32.1\%}} & \textit{\textcolor{ForestGreen}{+59.3\%}} \\
\cmidrule(lr){1-7}
\multirow{6}{*}{\shortstack[l]{\textbf{ChartSync}\\{\tiny 870 tasks}}}
 & Phi-3.5-V & \textbf{.819}\,\gain{+.056} & \textbf{.832}\,\gain{+.041} & .210\,\loss{+.057} & \textbf{.815}\,\gain{+.054} & \textbf{.668}\,\gain{+.022} \\
 & InternVL2.5 & \textbf{.838}\,\gain{+.119} & \textbf{.849}\,\gain{+.141} & .185\,\loss{+.031} & \textbf{.820}\,\gain{+.121} & \textbf{.743}\,\gain{+.140} \\
 & Qwen3.5-4B & .886\,\loss{-.002} & .879\,\tie{+.000} & .062\,\loss{+.048} & \textbf{.825}\,\gain{+.002} & .800\,\loss{-.011} \\
 & Qwen3.5-9B & \textbf{.927}\,\gain{+.011} & .918\,\loss{-.006} & .043\,\loss{+.017} & .853\,\loss{-.010} & .829\,\loss{-.025} \\
 & Granite & \textbf{.819}\,\gain{+.036} & \textbf{.810}\,\gain{+.017} & \textbf{.252}\,\gain{-.674} & \textbf{.768}\,\gain{+.022} & \textbf{.644}\,\gain{+.228} \\
 & \textit{rel.\ $\Delta$ of mean} & \textit{\textcolor{ForestGreen}{+5.4\%}} & \textit{\textcolor{ForestGreen}{+4.7\%}} & \textit{\textcolor{ForestGreen}{-40.9\%}} & \textit{\textcolor{ForestGreen}{+4.9\%}} & \textit{\textcolor{ForestGreen}{+10.6\%}} \\
\midrule
\multicolumn{2}{l}{\textit{improved in}} & \textit{19/20} & \textit{18/20} & \textit{10/20} & \textit{19/20} & \textit{17/20} \\
\bottomrule
\end{tabular}}
\vspace{-.2cm}
\end{table*}

Using the means over the 20 comparisons, SFT yields relative gains
of approximately 15.6\% in requirement recall, 19.8\% in coupled-update
recall, and 22.4\% in exact-edit rate.
These gains extend beyond explicit requirements to the necessary
coupled updates that maintain chart consistency, supporting the
dataset's utility on external benchmarks.
However, preservation improves less consistently. Although mean
gratuitous changes per task decrease by 10.8\% overall, they decrease
in only 10 of the 20 comparisons.
The decomposition exposes preservation regressions that would be
obscured by reporting only the overall improvement in exact-edit rate.

\textbf{Exactness filtering mainly improves coupled-update consistency.}
Fine-tuning on the released data raises coupled-update recall in 7 of 8 comparisons
and exact-edit rate in 6 of 8 relative to training before exactness filtering
(\Cref{sec:auditing}). Requirement recall and gratuitous changes remain close
(\Cref{app:filter-ablation}).

\FloatBarrier

\subsection{Diagnosing exact-edit performance}
\label{sec:eval-main}

Aggregate benchmark scores do not fully capture cross-modal edit grounding.
We compare existing benchmark measurements after fine-tuning with the completion, consistency,
and preservation outcomes measured by our protocol.

\textbf{CodeScore rises while executability declines.}
CodeScore improves in 19 of 20 model--benchmark comparisons, with a mean gain of
8.1 points, whereas ExecRate improves in only five and decreases by 3.0 percentage points
on average (\Cref{tab:benchmark-results}). CodeScore is an LLM-based assessment of
modification accuracy and code completeness, following the evaluation used by ChartEdit
\citep{zhao2025chartedit}.
The disagreement is largest on Chart2Code-L2, where CodeScore increases for every model
but ExecRate falls for four.
The disagreement motivates inspecting what was completed and what else changed
in both executable and non-executable outputs.

\begin{table*}[!ht]
\vspace{-.35cm}
\centering
\small
\setlength{\tabcolsep}{2.5pt}
\renewcommand{\arraystretch}{0.95}
\caption{Results under the inherited benchmark-facing metrics. Each cell reports the
fine-tuned value and its change from the base model.}
\label{tab:benchmark-results}
\begin{tabular}{l|rr|rr|rr|rr}
\toprule
Family & \multicolumn{2}{c|}{\textbf{ChartEdit}} & \multicolumn{2}{c|}{\textbf{Chart2Code{-L2}}} & \multicolumn{2}{c|}{\textbf{ChartM$^3$}} & \multicolumn{2}{c}{\textbf{ChartSync}} \\
\cmidrule(lr){2-3} \cmidrule(lr){4-5} \cmidrule(lr){6-7} \cmidrule(lr){8-9}
 & Exec. & Code. & Exec. & Code. & Exec. & Code. & Exec. & Code. \\
\midrule
Phi-3.5-V & \dd{0.860}{\tie{+0.000}} & \dd{\textbf{75.4}}{\gain{+11.6}} & \dd{0.334}{\loss{-0.066}} & \dd{\textbf{38.0}}{\gain{+10.4}} & \dd{0.650}{\loss{-0.010}} & \dd{\textbf{46.3}}{\gain{+11.2}} & \dd{0.930}{\loss{-0.010}} & \dd{\textbf{85.5}}{\gain{+1.4}} \\
InternVL2.5 & \dd{0.860}{\loss{-0.040}} & \dd{\textbf{71.6}}{\gain{+20.4}} & \dd{0.380}{\loss{-0.214}} & \dd{\textbf{32.6}}{\gain{+11.9}} & \dd{\textbf{0.740}}{\gain{+0.050}} & \dd{\textbf{43.3}}{\gain{+10.6}} & \dd{0.930}{\loss{-0.030}} & \dd{\textbf{88.2}}{\gain{+8.4}} \\
Qwen3.5-4B & \dd{\textbf{0.910}}{\gain{+0.010}} & \dd{\textbf{86.7}}{\gain{+3.6}} & \dd{\textbf{0.444}}{\gain{+0.030}} & \dd{\textbf{62.1}}{\gain{+1.6}} & \dd{\textbf{0.700}}{\gain{+0.040}} & \dd{\textbf{66.8}}{\gain{+13.1}} & \dd{0.980}{\loss{-0.010}} & \dd{\textbf{92.6}}{\gain{+0.3}} \\
Qwen3.5-9B & \dd{0.920}{\loss{-0.020}} & \dd{\textbf{89.8}}{\gain{+6.0}} & \dd{0.540}{\loss{-0.027}} & \dd{\textbf{73.4}}{\gain{+6.7}} & \dd{0.720}{\loss{-0.010}} & \dd{\textbf{74.8}}{\gain{+13.6}} & \dd{0.980}{\loss{-0.010}} & \dd{94.8}{\loss{-0.9}} \\
Granite & \dd{0.860}{\loss{-0.010}} & \dd{\textbf{79.7}}{\gain{+11.2}} & \dd{0.373}{\loss{-0.223}} & \dd{\textbf{43.1}}{\gain{+13.4}} & \dd{0.630}{\loss{-0.070}} & \dd{\textbf{48.4}}{\gain{+4.5}} & \dd{\textbf{0.900}}{\gain{+0.030}} & \dd{\textbf{83.4}}{\gain{+3.7}} \\
\bottomrule
\end{tabular}
\vspace{-.5cm}
\end{table*}

\begin{figure*}[!h]
  \centering
  \includegraphics[width=0.96\textwidth]{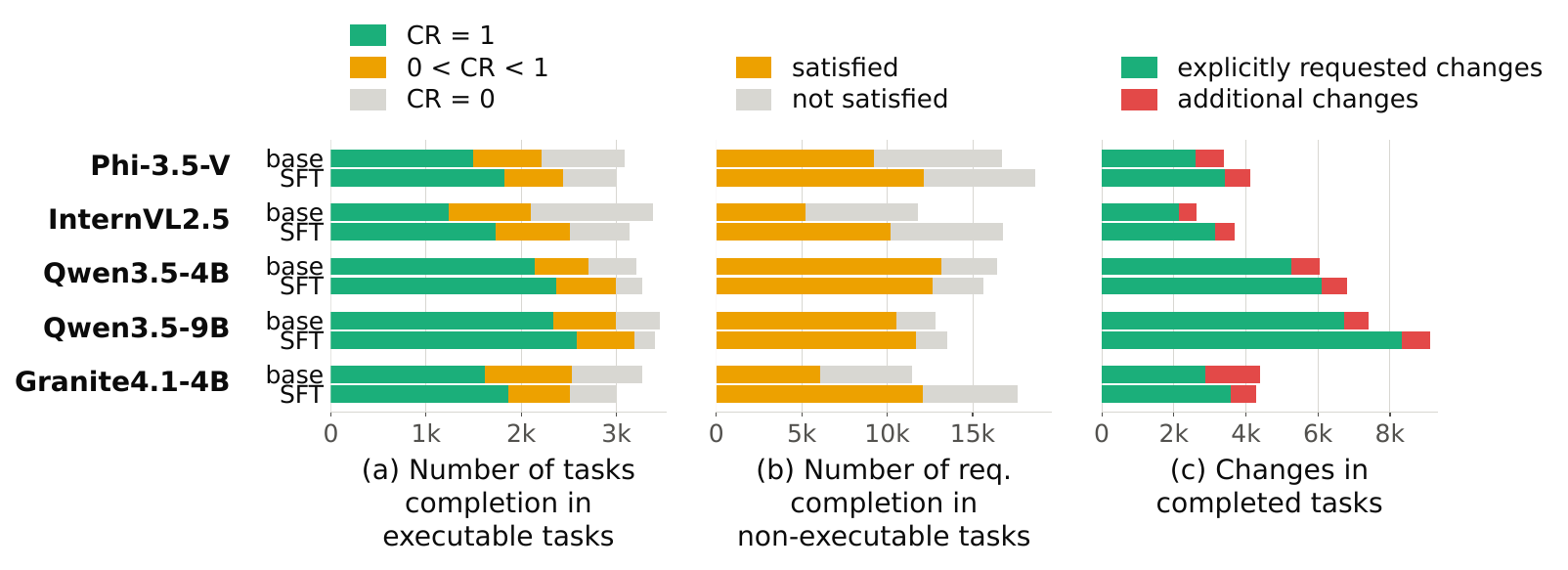}
  \vspace{-.4cm}
  \caption{Editing states pooled across benchmarks.
(a) Executable tasks grouped by completion ratio. (b) Requirements judged from code in non-executable outputs.
(c) Explicitly requested and additional changes (coupled and gratuitous) in tasks satisfying
all explicit requirements.}
  \label{fig:metric-blindspots}
\end{figure*}

\textbf{Completing explicit requirements is insufficient for exact editing.}
After SFT, all five models produce more executable outputs with full completion
($\mathrm{CR}=1$) and fewer with zero completion ($\mathrm{CR}=0$), although four produce
fewer executable outputs overall (\Cref{fig:metric-blindspots}(a)). For outputs that fail to
run, the number of requirements judged satisfied from code also rises for four of the
five models (\Cref{fig:metric-blindspots}(b)). 
Even outputs that satisfy every explicit requirement can still contain
additional changes, including necessary coupled updates and gratuitous
changes (\Cref{fig:metric-blindspots}(c)).

This breakdown helps interpret the divergence between CodeScore and
ExecRate, since requirement completion and execution capture
different aspects of an edit (\Cref{fig:metric-blindspots}(a)), and code-based checks can reveal partial
implementation even when a program fails to render (\Cref{fig:metric-blindspots}(b)).
\Cref{fig:metric-blindspots}(c) further illustrates why evaluating only explicit requirement
completion is insufficient. Exactness also depends on whether additional
changes are necessary or gratuitous and whether all necessary coupled
updates are satisfied. Our protocol therefore checks completion,
coupled-update consistency, and preservation separately, with execution
and rendering required for exact-edit success.

\begin{figure*}[!t]
\vspace{-.5cm}
  \centering
  \includegraphics[width=\linewidth]{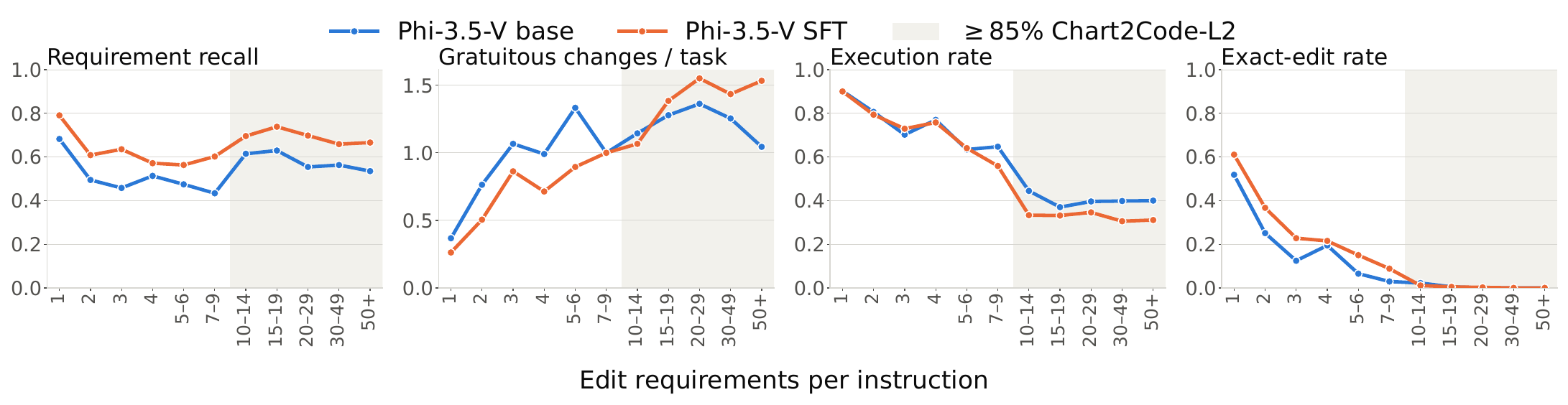}
  \vspace{-.5cm}
  \caption{Phi-3.5-Vision outcomes grouped by explicit requirement count across benchmarks. In shaded bins, at least 85\% of tasks come from Chart2Code-L2.}
  \label{fig:depth-exec}
  \vspace{-.2cm}
\end{figure*}

\textbf{The trade-off varies with edit difficulty.}
Preservation remains challenging both for complex requests and in settings with
little room for further improvement.
On Chart2Code-L2, four of five models introduce more gratuitous changes after SFT.
On ChartSync, Qwen3.5-4B and Qwen3.5-9B already achieve full-completion rates of
87.9\% and 92.4\% before SFT. Completion therefore changes little, while gratuitous edits
increase slightly.

For Phi-3.5-Vision, higher requirement recall after SFT coexists with more gratuitous
changes and lower executability on requests with many requirements. We measure \emph{edit
depth} by the number of explicit requirements. In \Cref{fig:depth-exec}, SFT improves
requirement recall across depth bins, but its advantage in reducing gratuitous changes
reverses for requests with at least 15 requirements. Exact-edit success remains near zero at deep depths. Deeper requests can involve more code and interacting components,
creating more opportunities for unrelated changes and execution errors. 
These results
highlight a trade-off, that encouraging models to complete more changes required by
complex requests can improve completion but increase execution risk, whereas conservative
editing by the base model may preserve executability while leaving requirements unmet.

\textbf{Execution errors.} Most of the increase in execution failures after SFT falls within exception classes already observed in the base models (\Cref{app:exec-error-audit}).

\subsection{Reliability of the evaluation}
\label{sec:human-transfer}

Evaluation protocol judgments agree with human assessments on 93\% of the audited outputs,
and both evaluations find positive fine-tuning gains on all four audited benchmark
subsets. The audit covers blinded outputs from 200 paired tasks in which both programs
render (\Cref{app:human-transfer}). A positive human exact-edit judgment requires every
listed requirement to be satisfied and no gratuitous change. The protocol and human
evaluations estimate aggregated gains of 14.0 and 14.5 percentage points, respectively, supporting agreement on the direction of the gains.

Some disagreements with human judgments arise from render-time layout effects, such as overlapping elements.
These cases reveal a limitation of our code-only protocol, that code permits
checks of object properties and dependencies even for non-executable
outputs, but does not directly show their rendered layout.
Chart evidence could complement these checks by revealing layout errors
that violate explicit requirements, necessary coupled updates, or
preservation. But the general aesthetic quality remains outside of the evaluation scope.

%% file: 6-conclusion.tex
\section{Conclusion}
\label{sec:conclusion}
\sysname{} provides a grammar-derived dataset and a reference-free protocol for exact program-grounded chart editing. Operation templates with source-program checks give systematic coverage across chart types and libraries, and requirement-level audits catch incomplete edits during construction. The protocol separately scores explicit requirements, coupled updates, and gratuitous changes, and agrees with human exact-edit judgement on 93\% of audited outputs. Across five models and four external benchmarks, fine-tuning improves requirement recall in 19 of 20 comparisons and exact-edit rate in 17 but reduces gratuitous changes in only 10, leaving preservation on complex requests as the main open challenge.

\newpage
\subsection*{AI Use Statement}

We used generative AI to generate edited chart programs and assist automated validation,
feedback-based repair, and auditing; LLM also serves as the judge in our
evaluation protocol and benchmark-facing metrics. Additionally, we used generative AI tools to edit the manuscript for readability and to review manuscript drafts. We have reviewed all AI-assisted work: dataset quality was checked through execution checks, cross-model auditing, and human validation of sampled records (\Cref{sec:release}), the automatic judge was compared with human judgments (\Cref{sec:human-transfer}), and the authors checked all manuscript text. We take responsibility for the final content of this work, including text, claims, and artifacts produced with the aid of generative AI.

\subsection*{Ethics Statement}

The dataset is derived from ChartNet's publicly available synthetic chart programs~\citep{kondic2026chartnet}. Reuse of the source material remains subject to its license and
attribution requirements. The four external
benchmarks are used only for evaluation under their original terms and are not redistributed. All \sysname{}'s
charts are synthetic and contain no personal information. Human audits assess editing correctness on sampled records and
model outputs. Since automatic editing can alter data, omit requested updates, or introduce
unrelated changes, generated charts require verification before use.

\subsection*{Reproducibility Statement}
The appendix documents the operation inventory, grounding procedure, and construction
prompts (\Cref{app:construction}), training and inference settings
(\Cref{app:training,app:config}), and the evaluation-protocol prompts with a worked example
(\Cref{app:exact-prompts,app:concepts}). Complete external-benchmark results, protocol diagnostics,
and the exactness-filtering ablation are provided in \Cref{app:eval-supplement}. The release schema
distinguishes distributed records from internal collection logs (\Cref{app:schema}), while the
human-audit descriptions specify the sampled populations and correctness criteria
(\Cref{app:release-audit,app:human-transfer}). We will release the dataset, the evaluation code, and the full prompts upon publication.

%% file: 99-appendix.tex
%
%
%
%
%
%

\lstdefinestyle{prompt}{
  basicstyle=\ttfamily\scriptsize,
  breaklines=true,
  breakindent=0pt,
  columns=fullflexible,
  frame=single,
  framesep=4pt,
  rulecolor=\color{gray!50},
  backgroundcolor=\color{gray!5},
  xleftmargin=2pt,
  xrightmargin=2pt,
}

\section{Extended Related Work}
\label{app:extended-related}

\textbf{Chart understanding and chart-to-code tasks.}
Related work covers chart understanding and generation
\citep{masry-etal-2022-chartqa,wu2025plot2code,yang2025chartmimic,liu2023matcha,masry2023unichart,han2023chartllama,xia2025chartx,masry2025chartgemma,zhao2025chartcoder,he2026chart},
code generation and editing \citep{chen2021evaluating,cassano2023can,jimenez2024swe},
and instruction-guided image editing
\citep{brooks2023instructpix2pix,zhang2023magicbrush,sheynin2024emu,wang2023imagen}.
Our setting requires implementing a visual edit request in an existing plotting program
while preserving unrelated chart content. This requires grounding each visual request in
both the chart's components and their code implementation.

\textbf{Chart editing and coverage.}
Chart-editing resources cover program revision
\citep{zhao2025chartedit,tang2026charts,kapadnis2026charteditbench}, multimodal instructions
\citep{yang2025chartm3,chen2026charteditor}, and image-space edits
\citep{li2026charts,yu2026chartsync}. Our evaluation provides the source chart and program
and requires an edited program (\Cref{app:config}). For construction, we use the grammar
of graphics \citep{wilkinson2011grammar,wickham2010layered,satyanarayan2016vega} to organize
semantic object--action templates with source-specific applicability checks across chart
types and libraries. Operation categories and interaction formats describe different
aspects of coverage. \sysname{} therefore emphasizes source-grounded applicability and
reuse of semantic edit definitions across plotting libraries.

\textbf{Evaluation.}
Atomic decomposition and instruction-aware visual evaluation provide related
foundations \citep{cho2024davidsonian,ku2024imagenhub,ku2024viescore}.
In chart editing, ChartM$^3$ and FigEdit measure instruction compliance and preservation
\citep{yang2025chartm3,li2026charts}; ChartEditBench uses assertions for programmatic
instructions \citep{kapadnis2026charteditbench}; and ChartSync checks text-to-geometry
dependencies \citep{yu2026chartsync}. Our protocol jointly checks explicit requirements,
source-derived coupled states, and unrelated changes in the edited program. This separates
completion, consistency, and preservation without a reference edit, with successful
execution and rendering required for exact-edit success (\Cref{sec:exact-edit-protocol}).

\textbf{Judge reliability.}
LLM judges can exhibit position and self-preference biases
\citep{wang2024large,panickssery2024llm}. Following work on alignment with human judgments
\citep{liu2023g,zheng2023judging}, we report cross-model construction audits and human
agreement (\Cref{sec:auditing,sec:human-transfer}). Our validation targets edit-specific
requirement satisfaction and preservation in chart programs, rather than general response quality.

\FloatBarrier
\section{Dataset Construction and Prompts}
\label{app:construction}
\label{app:prompts}

This section follows the construction pipeline of \Cref{sec:taxonomy}. Prompt descriptions are provided at the stage where they are used.

\subsection{Taxonomy scope and library realizations}
\label{app:grammar}

\Cref{tab:grammar} lists the program objects through which each category is
realized. The same object can serve several categories; the request determines the category.

\begin{table}[!ht]
\centering
\small
\caption{{Semantic objects of the twelve edit categories.}}
\label{tab:grammar}
\begin{tabular}{lp{0.74\linewidth}}
\toprule
Category & {Semantic objects} \\
\midrule
data & dataset, field, record, series, category \\
transform & aggregation, bin, stack, derived metric, sort, window \\
encoding & position, colour, size, shape, opacity, and text channels \\
layer & primary and overlay layers, error bars, reference line or band, secondary axis \\
mark & line, point, bar, area, arc, box, violin, cell \\
position & margins, anchor, z-order, aspect ratio, legend and annotation placement \\
scale & position, colour, size, and opacity scales, normalization \\
guide & axis, ticks, gridlines, legend, colorbar \\
coordinate & Cartesian axes, polar coordinates, spines \\
facet & subplot grid, facet layout, shared axes, panel title \\
annotation & text, arrow, data label, shape, image \\
style & figure, plot area, title, axis label, theme, palette \\
\bottomrule
\end{tabular}
\end{table}

\textbf{Extending the inventory.}
A new edit type needs a target, action template, and applicability checks. A new chart type
needs compatibility entries; a new library needs source-program bindings and validation.
The twelve augmentation templates extend existing categories under these same requirements.

\subsection{Operation inventory}
\label{app:taxonomy}

The inventory contains \NEditsAll{} object--action templates across \NCategories{} categories.
The complete main-template list and compatibility rows are stored in
\texttt{edit\_matrix.json}. {\Cref{tab:grammar} lists target objects, and
\Cref{tab:percat-release} gives released record counts by category.}

\subsection{Grounded instruction materialization}
\label{app:grounding}

\Cref{tab:grounding} lists recorded template--instruction pairs. Field placeholders are
filled from source-program values, while arguments such as a new grid size specify a requested
state.
\begin{table}[ht]
\centering
\small
\caption{Template $\rightarrow$ grounded instruction examples.}
\label{tab:grounding}
\resizebox{\linewidth}{!}{
\begin{tabular}{llp{9.0cm}}
\toprule
Edit & Chart & Template $\rightarrow$ realized instruction \\
\midrule
E026 & Bubble (seaborn)
  & \texttt{Add subgroup by \{subgroup\_field\} within \{primary\_group\_field\}} \newline
    $\rightarrow$ ``For the subgroup, add subgroup by \texttt{`Crop`} within
    \texttt{`Region`}.'' \\
\addlinespace
E311 & Bubble (seaborn)
  & \texttt{Map \{facet\_row\_field\} to rows and \{facet\_column\_field\} to columns} \newline
    $\rightarrow$ ``For the facet layout, map \texttt{`Region`} to rows and \texttt{`Crop`}
    to columns.'' \\
\addlinespace
E308 & Pie (matplotlib)
  & \texttt{Set the multi-facet grid to \{facet\_rows\} rows and \{facet\_columns\} columns}
    \newline
    $\rightarrow$ ``For the subplot grid, set the multi-facet grid to 2 rows and 2 columns.'' \\
\addlinespace
E100 & Bubble (plotly)
  & \texttt{Remove facet mapping for \{facet\_field\}} \newline
    $\rightarrow$ ``For the facet channel, remove facet mapping for \texttt{`Gender`}.'' \\
\bottomrule
\end{tabular}
}
\end{table}

\paragraph{Geometry-bound arguments.}
Some arguments bind against the chart's rendered geometry rather than its data schema. For
example, a requested figure width is floored against the length of the chart's own title, and
legend-anchoring instructions name \texttt{bbox\_to\_anchor} explicitly.

\subsection{Program generation: editor prompt}
\label{app:prompt-editor}

The editor and first-pass judge use \texttt{gemma-4-31B-it} served locally. The editor
receives this system prompt, then the instruction and the complete source program, and returns
a full-file rewrite, so that a patch that fails to apply cannot be confused with an editing
failure.

\begin{lstlisting}[style=prompt]
You are a chart-editing engine. Follow the edit request exactly.

Respond with ONLY the complete edited Python program in one ```python code block.

This is a full-file rewrite, not a diff or patch. Do not omit unchanged code,
abbreviate with ..., or use comments like "# unchanged" or "# rest of code stays
the same". Write every line of the runnable program in full.

Make exactly the requested edit and no additional edit. Preserve all unrelated
chart data, encodings, labels, legends, annotations, scales, layout, styling,
chart type, and save behavior unless the instruction explicitly asks to change
them.
\end{lstlisting}

\subsection{Initial collection validation}
\label{app:prompt-judge}

This first-pass collection screen uses code and image evidence. It is distinct from the
code-only exact-edit audit in \Cref{app:exact-prompts}. The judge receives the instruction,
named target, source and edited programs, code diff, and before/after renders. The condensed
prompt below describes this screen.

\begin{lstlisting}[style=prompt]
You are a meticulous chart-edit reviewer. You are given:
1. The edit instruction and its named target object
2. BEFORE and AFTER chart images
3. The SOURCE program
4. The COMPLETE AFTER program
5. The unified code diff

Anchor every judgment on the NAMED TARGET OBJECT. First establish what happened to
that object specifically -- compare its appearance, content and position between
BEFORE and AFTER, and check the diff for a change that actually targets it. Only then
consider the rest of the chart.

Judge four independent questions:
- applied: Was the exact requested edit carried out ON THAT TARGET OBJECT? Use the
  diff and the complete program to confirm the mechanism really lands (a value set but
  never used, or overwritten by a later line, is not applied; a comment describing an
  effect is not an implementation). Use the images to confirm it landed on the RIGHT
  object -- a change that hits a different object than the one named is applied:false,
  even if the code looks correct. Visually subtle changes (a linewidth, a small margin,
  a few points of font size) still count as applied when the code unambiguously
  implements them.
- no_additional_change: Was everything else left unchanged, with no extra edit?
- position_preserved: Did every object (marks, axes, legend, colorbar, annotations,
  labels, subplots/facets) keep its correct position? If the instruction itself asked
  to move something, judge only whether THAT move landed and nothing else shifted.
- no_object_removed: Did every visual object that existed in BEFORE, and that the instruction
  did NOT ask to remove, still exist in AFTER? Inventory the BEFORE image's furniture --
  legend, colorbar, title, axis labels, tick labels, pre-existing data labels,
  annotations, gridlines, error bars, reference lines, each data series -- and confirm
  each survives. A frequent failure is an edit that lands correctly while silently
  destroying something unrelated (adding a data label but dropping the legend,
  replotting a series so earlier labels vanish). If the instruction explicitly asked to
  remove that object, this stays true.

Collateral damage ELSEWHERE does not make applied false -- it belongs in the other
three fields. Damage to the TARGET ITSELF does make applied false, per the rule above.

Count changes to data, labels, legends, colorbars, annotations, marks, encodings, axes,
scales, layout, styling, sorting, filtering, faceting, chart type, or save behavior as
additional changes unless explicitly requested.

The reason MUST state what happened to the named target object specifically -- not a
generic statement that the chart changed.

Respond with ONLY one JSON object:
{"applied": true/false, "no_additional_change": true/false,
"position_preserved": true/false, "no_object_removed": true/false,
"reason": "what happened to the target object, in one or two short sentences"}
\end{lstlisting}

The main collection stores \texttt{render\_ok} and the four judge flags. Applied and strict
success share the all-attempt denominator, so their difference counts applied-but-not-strict
attempts.

\subsection{Feedback repair prompt}
\label{app:prompt-repair}

A repair attempt is issued when a candidate fails to render or is judged not applied, and
carries the traceback or the judge's rationale. The release was cleaned in three passes, each
allowing up to two further attempts.

\begin{lstlisting}[style=prompt]
You are a chart-editing engine. You previously attempted to apply an edit instruction
to a Python plotting program, but the result had a problem -- you are given the
ORIGINAL instruction, YOUR PREVIOUS attempt's code, and FEEDBACK describing exactly
what went wrong (either a render error/traceback, or a reviewer's note that the
instruction wasn't actually carried out). Produce a corrected version of the program
that still applies the original instruction and fixes the specific problem in the
feedback. Do not reintroduce the same mistake, and do not undo the original edit while
fixing the problem.
This is NOT a diff or patch task -- respond with ONLY the full corrected program, every
line of it, in a single ```python code block, ready to run as-is. Never abbreviate or
omit unchanged code with placeholders like '...', '# unchanged', or similar -- write out
every line in full, exactly as if you were submitting the complete file.
\end{lstlisting}

\subsection{Compound auditing and post-processing}
\label{app:compound}

\paragraph{Audit stages and counts.}
The atomic audit starts from the 5{,}759 compound records of the preceding release. Of 710
defective records it retried, 472 were recovered. Unstable tier-3 records were excluded after a
pilot on 60 of them recovered only 40\%. In total 1{,}240 records were removed (997 tier-3, 218
tier-2, and 25 templated), leaving \NReleasedCompound{} compound records. The 34.6\% omission
rate in \Cref{sec:auditing} is measured on the 4{,}611 tier-2 and tier-3 records that carried an
initial verdict, of which 1{,}596 had been marked applied although at least one requirement was
not met.

The quarantine audit groups defective instructions into seven recurring classes. The most
common failures are cross-field: incompatible operands,
generated field names that do not match the requested operation, numeric formats that do not
match the field type, degenerate target choices, and fabricated values. Defective records are
excluded from the release.

\FloatBarrier
\section{Dataset Characterization and Release Validation}
\label{app:dataset-quality}

This section gives coverage details, collection outcomes, and the accepted-set audit
summarized in \Cref{sec:release}.

\subsection{Coverage notes}
\label{app:coverage}

Each record retains its source \texttt{pool\_idx}. When the source pool lacks a required
object, such as an annotation arrow or reference line, an auxiliary source chart is derived from
an existing program and the precondition is checked again.

A \emph{compatibility cell} is an (edit type, chart type) pair. The initial compatibility
matrix contains \NCells{} eligible cells. Collection uses 25 successful records per cell
as a sampling target; this is not a guarantee that every cell reaches that count, or that
all collected successes survive release filtering.

The released manifest, which also includes later templates and compound records, contains
4{,}619 populated (template, chart type) pairs; 677 hold at least 25 records. These released
counts are separate from the initial matrix and its collection target. At the coarser level
used in \Cref{fig:coverage}, the twelve categories plus the compound stratum form 258 populated
(category or stratum, chart type) bins, 256 of them with at least 25 records. Coverage is therefore
non-uniform, and population of a cell should not be read as attainment of the sampling target.

\subsection{Released dataset composition}

{\Cref{tab:percat-release} reports the final training and test records by category.
Compound requests combine operations from the twelve categories and are listed separately.}

\begin{table}[!ht]
\centering
\small
\caption{Per-category composition of the released dataset. Compound records are training-only.}
\label{tab:percat-release}
\begin{tabular}{lrrr}
\toprule
Category & Train & Test & Total \\
\midrule
guide       & 14{,}356 & 253 & 14{,}609 \\
data        & 11{,}223 & 218 & 11{,}441 \\
style       & 10{,}994 & 252 & 11{,}246 \\
position    &  8{,}521 & 223 &  8{,}744 \\
annotation  &  8{,}385 & 187 &  8{,}572 \\
transform   &  7{,}171 & 177 &  7{,}348 \\
layer       &  5{,}583 &  90 &  5{,}673 \\
encoding    &  4{,}950 &  83 &  5{,}033 \\
scale       &  4{,}933 & 120 &  5{,}053 \\
facet       &  4{,}422 &  97 &  4{,}519 \\
compound    &  4{,}519 &   0 &  4{,}519 \\
mark        &  2{,}971 &  69 &  3{,}040 \\
coordinate  &  2{,}561 &  80 &  2{,}641 \\
\midrule
\textbf{All} & \textbf{90{,}589} & \textbf{1{,}849} & \textbf{92{,}438} \\
\bottomrule
\end{tabular}
\end{table}

The \NChartTypes{} chart types are: Area Chart, Bar Chart, Box Plot, Bubble Chart, Funnel
Chart, Heatmap, Histogram, Kernel Density Estimate Plot, Line Chart, Multi-Axes Chart, Pie
Chart, Radar Chart, Ring Chart, Rose Chart, Scatter Plot, Stem Plot, Swarm Plot, Tornado
Chart, Treemap, and Violin Plot.

\FloatBarrier
\subsection{Dataset characterization and human assessment}
\label{app:release-audit}

\begin{figure}[!ht]
  \centering
  \includegraphics[width=.8\linewidth]{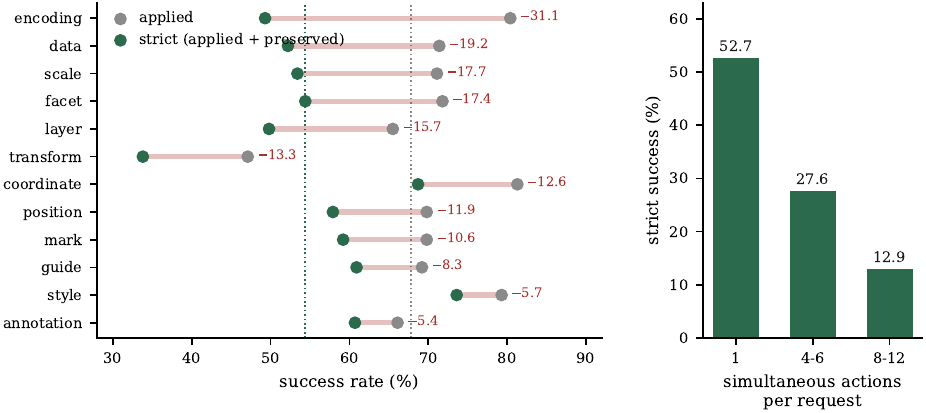}
  \caption{Preservation is a second axis of difficulty, not a proxy for the first. (a) The bar
  joining the two markers is the applied/strict gap; dotted lines are the earlier export means.
  (b) Strict success by collection subset. Tier~2 contains 4--6 actions and tier~3 contains
  8--12.}
  \label{fig:findings}
\end{figure}

\Cref{fig:findings} plots (a) applied and strict success by category and (b) strict success
by collection subset, from the collection-stage export.

\textbf{Human assessment of accepted training records.}
{We audit a stratified sample of 500 accepted records: 300 from the
86{,}070 single-action records and 200 from the 4{,}519 compound records. Each record is
executed and inspected against its source.}

\textbf{Task-level correctness} requires successful execution, satisfaction of every
explicit requirement, and preservation of unrelated chart content. The overall estimate
weights the two subsets by their proportions in the training set. \textbf{Requirement-level correctness}
measures individual requested changes within the 200 audited compound records; it does
not require the entire record to pass. \Cref{tab:release-audit} reports both levels.

\begin{table}[t]
\centering
\caption{Human validation of accepted training records. Panel~(a) reports task-level
correctness by subset and the training-set-weighted estimate. Panel~(b) reports correctness
of individual requirements in the same 200 compound records.}
\label{tab:release-audit}
\small
\begin{tabular}{lrrrr}
\toprule
Subset & Audited units & Correct & Correctness & 95\% CI \\
\midrule
\multicolumn{5}{l}{\textbf{(a) Task level: records}} \\
Single-action           & 300 & 255 & 85.0\% & [80.5, 88.6] \\
Compound                & 200 & 139 & 69.5\% & [62.8, 75.5] \\
Weighted & 500 & --- & \textbf{84.2\%} & [80.4, 88.1] \\
\midrule
\multicolumn{5}{l}{\textbf{(b) Requirement level: compound requirements}} \\
Compound                & 1{,}071 & --- & \textbf{95.6\%} & [94.2, 96.7] \\
\bottomrule
\end{tabular}
\end{table}

\FloatBarrier
\section{Training and Evaluation Details}
\label{app:metrics}
\label{app:additional-eval}

This section gives training and inference settings, metric conventions, and the code-only
exact-edit protocol prompts.

\subsection{Training configuration}
\label{app:training}

All five fine-tuned checkpoints in \Cref{sec:finetune} are full-parameter supervised
fine-tunes of the corresponding base model. Each model is trained for one epoch on the released training split
(\NReleasedTrain{} records), starting from the public base checkpoint. Every run uses learning rate $1\times10^{-5}$
with a cosine schedule, bfloat16 precision, weight decay 0, gradient clipping at 1.0, and
gradient checkpointing. We evaluate the weights saved at
the end of the epoch.
\Cref{tab:train-config} lists the settings that differ across model families.

\begin{table}[h]
\centering
\small
\setlength{\tabcolsep}{3pt}
\caption{Per-model fine-tuning settings. Effective batch is per-device batch $\times$ gradient
accumulation $\times$ number of GPUs. Settings shared by all runs are given in the text.}
\label{tab:train-config}
\resizebox{\linewidth}{!}{
\begin{tabular}{lccccc}
\toprule
 & Phi-3.5-V & InternVL2.5-8B & Qwen3.5-4B & Qwen3.5-9B & Granite-4.1-4B \\
\midrule
Effective batch        & $1\times16\times2=32$ & $1\times8\times4=32$ & $4\times4\times4=64$ & $4\times2\times8=64$ & $1\times16\times4=64$ \\
Optimizer steps        & 2{,}830 & 2{,}805 & 1{,}416 & 1{,}416 & 1{,}416 \\
Warmup steps           & 20 & 20 & 60 & 60 & 60 \\
Optimizer              & AdamW (8-bit) & AdamW (8-bit) & AdamW (8-bit) & AdamW & AdamW (8-bit) \\
Max sequence length    & 8{,}192 & 6{,}144 & 8{,}192 & 8{,}192 & 8{,}192 \\
Image input            & 1 crop & $\le$6 tiles + thumbnail & $\le$1{,}280 visual tokens & $\le$1{,}280 visual tokens & $384\times384$ \\
Training records       & 90{,}589 & 89{,}746$^{\dagger}$ & 90{,}589 & 90{,}589 & 90{,}589 \\
Parallelism            & DDP & ZeRO-3 & ZeRO-3 & ZeRO-3 & ZeRO-2 \\
Hardware               & 2$\times$A100 40\,GB & 4$\times$A100 40\,GB & 4$\times$A100 40\,GB & 8$\times$80\,GB GPUs & 4$\times$L40S 48\,GB \\
\bottomrule
\end{tabular}}
\\[2pt]
{\footnotesize $^{\dagger}$843 records whose tokenized length exceeds the 6{,}144-token limit
are dropped rather than truncated.}
\end{table}

\FloatBarrier
\subsection{Inference and evaluation configuration}
\label{app:config}

Evaluation uses the model families listed in \Cref{sec:eval-setup}. Each editing model receives the
rendered source chart, complete source program, and text edit instruction and returns a full
edited program. The automatic exact-edit judge receives code and textual task information
only; rendered images are not inputs to that judge.
All four benchmarks provide source code for this evaluation; metrics requiring visual
pointers are not used. The evaluated task counts are 1{,}405 for ChartEdit, 1{,}010 for
Chart2Code-L2, 1{,}000 for ChartM$^3$, and 870 for ChartSync.

Base and fine-tuned models are served with vLLM~\cite{kwon2023efficient} (maximum model length 32{,}768) under the same
settings: greedy decoding (temperature 0), at most 6{,}144 new tokens, and no repetition
penalty. Every generated program is executed in a single rendering environment (Python 3.12, Matplotlib
3.11.1, Plotly 6.9.0) with a 60-second timeout per program.
The exact-edit judge settings
are given in \Cref{app:exact-prompts}.
\Cref{tab:metric-inventory} lists the metrics and their provenance.

\begin{table*}[t]
\centering
\small
\setlength{\tabcolsep}{3pt}
\renewcommand{\arraystretch}{1.08}
\caption{Metric inventory and provenance.}
\label{tab:metric-inventory}
\resizebox{\linewidth}{!}{
\begin{tabular}{p{2.8cm}p{2.4cm}p{7.6cm}}
\toprule
Metric & Source & Meaning / use \\
\midrule
ExecRate & Mechanical & Fraction of all tasks whose generated program executes and renders. \\
CodeScore & LLM judge & Continuous editing-quality score over the generated program; can score non-running programs. \\
ChartScore & LLM judge & Overall rendered-chart quality; not target-edit success. \\
AppliedRate & LLM judge & Whether the requested edit reached the named target. \\
EditFidelity & LLM judge & Applied edit with no unrequested change, unintended reposition, or destroyed object. \\
Requirement recall (CR$_\mu$) & GT-free judge & Micro-aggregated fraction of explicit requirements satisfied across all tasks. \\
Full completion & GT-free judge & Fraction of tasks satisfying every explicit requirement, independently of execution. \\
Gratuitous changes & GT-free judge & Unrequested changes that are not unavoidable consequences of requested edits, averaged per task. \\
Coupled-update recall & GT-free judge & Fraction of necessary updates satisfied, using one fixed source-derived inventory per task. \\
Missed coupled updates & GT-free judge & Unmet fixed-inventory end states: omitted or incorrectly implemented, including when the forcing requirement is unmet. \\
Exact-edit rate & Code judge + execution & Successful execution/rendering and a complete verdict with all requirements satisfied and neither deviation. \\
TESR / VLCS / BFS & ChartSync & Native text-edit, geometry, and whole-output fidelity signals. \\
\bottomrule
\end{tabular}}
\end{table*}

\subsection{Exact-edit protocol prompts}
\label{app:exact-prompts}
The four listings below summarize the code-only exact-edit protocol. They describe its
inputs, decision rules, and JSON schemas. The judge receives the instruction-derived requirements, source and edited
programs, and, for coupled checks, the fixed inventory of necessary updates. It receives no
rendered images. Execution and rendering are checked separately to determine the executable
success.

External-benchmark evaluation uses \texttt{gemma-4-31B-it} with greedy decoding
(\texttt{temperature}~0, \texttt{seed}~23) and JSON output. Stages A1 and A2 run once per task before any model output is seen.
The construction cross-audit uses \texttt{Qwen3.8-27B} with the same code-only protocol,
decision rules, and output schemas. Changing the judge model does not introduce a separate
definition of exactness. The earlier four-flag, code-and-image screen is documented in
\Cref{app:prompt-judge}.

\paragraph{Scope and correctness are separate checks.}
\label{app:deviation-boundary}
A coupled update is a required change to a source-dependent state. Its correctness is
checked against the intended end state, including when the program updates it automatically.
An omitted or incorrectly implemented update fails this check. A gratuitous change is an
actual modification outside the scope of the explicit requirements and necessary coupled
updates. Failure of an in-scope update is not, by itself, evidence of an additional
out-of-scope modification. An output can fail both checks when it also changes unrelated
content; the coupled inventory and extra-change list provide separate evidence for these
judgments. Properties that merely need to remain unchanged belong to preservation and are
not added to the coupled inventory solely because they existed in the source.

The reported coupled-update count measures unmet required end states, covering both omission
and incorrect implementation. It does not distinguish these two subtypes. When the forcing
requirement is not implemented, its coupled end state is also marked unmet under the fixed
inventory; this score must therefore be interpreted alongside explicit requirement completion.

\subsubsection{Stage A1: requirement decomposition}
\label{app:prompt-decompose-req}

\begin{lstlisting}[style=prompt]
Decompose the chart-editing instruction into atomic requirements.
Each requirement is one independently checkable change, stated as a visible property
of the edited chart. Split compound requests into separate items, keep the
instruction's wording, and do not add requirements it does not state. Omit anything
that is not visible in the rendered chart (variable names, comments, file paths).

INSTRUCTION: {instruction}

Return JSON: {"requirements": ["<requirement>", ...]}
\end{lstlisting}

\subsubsection{Stage A2: coupled-update inventory}
\label{app:prompt-decompose-coupled}

\begin{lstlisting}[style=prompt]
List every coupled update: a visible change that the requirements do not state but that
any correct implementation must make, because an element of the SOURCE program depends
on something a requirement changes. Typical cases:
  - a legend entry, label, title, annotation or tick label that refers to changed data
  - an axis limit, tick set or colorbar range that no longer fits the changed data
  - arrays that must keep matching lengths, or style lists sized to the number of series
  - positions or offsets computed from the number of groups or the changed values
Name the source element behind each item and the requirement that forces it. List an
item even if it would update automatically. Do not restate a requirement. Return an
empty list if nothing depends on the requested changes. Do not add unrelated properties
that merely need to remain unchanged; preserving them is a preservation check, not a
necessary coupled update.

SOURCE: {source}
REQUIREMENTS: {requirements}

Return JSON: {"coupled": [{"update": "<required end state>", "element": "<source element>",
                           "forced_by": <requirement number>}, ...]}
\end{lstlisting}

\subsubsection{Requirement satisfaction and gratuitous changes}
\label{app:prompt-judge-exact}

\begin{lstlisting}[style=prompt]
Audit the SOURCE and EDITED programs against the fixed requirements. No chart images
are provided. Infer chart properties from the plotting code and its data. A requirement
is satisfied only when the edited program implements the requested state on its target;
a comment, unused assignment, or overwritten setting is not sufficient evidence. Judge
these component properties independently of the program's overall execution outcome.
1. For each requirement, decide whether it is satisfied.
2. List additional chart-affecting changes evident in the programs and classify their scope:
   - "necessary": a companion change required by an explicit requirement;
   - "gratuitous": a change outside the explicit requirements and required companions.
   Whether a necessary companion was implemented correctly is checked separately. Do not
   infer a gratuitous change solely from an omitted or incorrect in-scope update. If the
   output also changes an unrelated property, list that modification as gratuitous.
   If unsure whether an additional modification is in scope, classify it as gratuitous.
Ignore code-only differences that cannot affect the chart: formatting, comments, output
file paths, and equivalent rewrites.

SOURCE: {source}
REQUIREMENTS: {requirements}
EDITED: {edited}

Return JSON: {"requirements": [{"i": 1, "satisfied": true|false}, ...],
              "extra_changes": [{"what": "<change>", "kind": "necessary"|"gratuitous"}, ...]}
\end{lstlisting}

\subsubsection{Coupled-update satisfaction}
\label{app:prompt-judge-coupled}

\begin{lstlisting}[style=prompt]
Audit the SOURCE and EDITED programs against the fixed coupled-update inventory.
No chart images are provided. Infer the required end states from code and data.
For each numbered coupled update (each forced by one requirement), decide:
  trigger_made: does the edit make, fully or partly, the change of the requirement that
                forces it?
  satisfied:    does the EDITED program implement the required end state, whether through
                an explicit edit or automatic synchronization? An omitted or incorrect
                implementation is false. If the trigger was not made, answer false.

SOURCE: {source}
REQUIREMENTS: {requirements}
COUPLED UPDATES: {coupled}
EDITED: {edited}

Return JSON: {"coupled": [{"k": 1, "trigger_made": true|false, "satisfied": true|false}, ...]}
\end{lstlisting}

\tcbset{
  conceptcard/.style={
    enhanced, breakable, colback=blue!2!white, colframe=blue!45!black,
    fonttitle=\bfseries\small, boxrule=0.6pt, arc=2pt, left=5pt, right=5pt,
    top=4pt, bottom=4pt, title={#1}},
  failcard/.style={
    enhanced, breakable, colback=red!2!white, colframe=red!60!black,
    fonttitle=\bfseries\small, boxrule=0.6pt, arc=2pt, left=5pt, right=5pt,
    top=4pt, bottom=4pt, title={#1}},
  passcard/.style={
    enhanced, breakable, colback=green!2!white, colframe=green!45!black,
    fonttitle=\bfseries\small, boxrule=0.6pt, arc=2pt, left=5pt, right=5pt,
    top=4pt, bottom=4pt, title={#1}},
}

\subsection{Examples of requirements, coupled updates, and gratuitous changes}
\label{app:concepts}

The following ChartEdit examples illustrate explicit requirements, coupled updates, and
gratuitous changes. Task~451 is presented as a source-based illustration of a missed versus
completed legend update. Tasks~1226 and~585 show fine-tuned and base Phi-3.5-V outputs,
respectively.

\begin{tcolorbox}[failcard={Gratuitous change: every requirement met, plus an unrequested edit},breakable=false]
\textbf{ChartEdit task 1226.} \emph{``Modify the 1.8 value of the `MRR' data series from 0.256 to
0.260, ensuring the marker remains square and orange.''} Three requirements (the value at 1.8 is
0.260, and its marker is square and orange) define the requested edit.

\smallskip
\begin{minipage}[t]{0.38\linewidth}
\includegraphics[width=\linewidth]{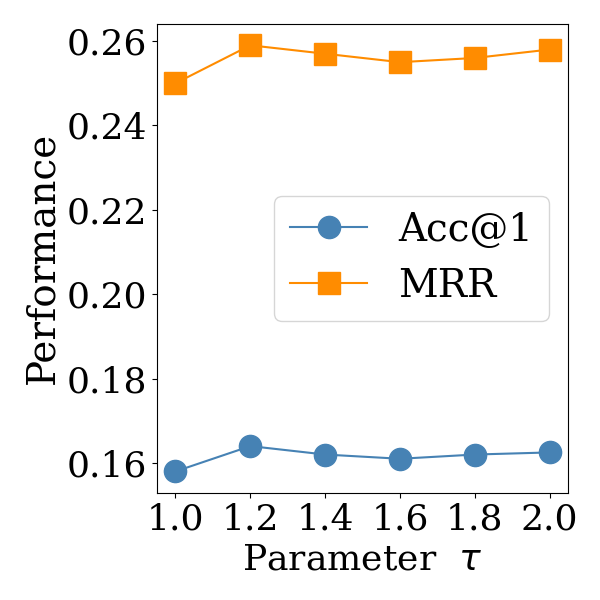}
\\{\footnotesize source}
\end{minipage}\hfill
\begin{minipage}[t]{0.38\linewidth}
\includegraphics[width=\linewidth]{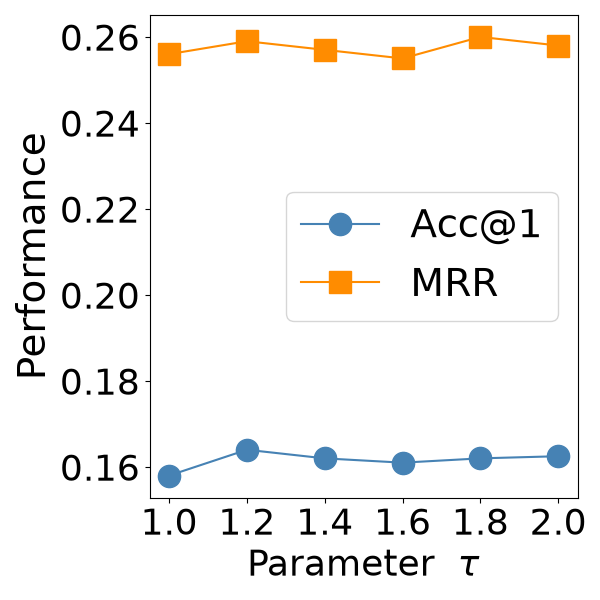}
\\{\footnotesize fine-tuned output}
\end{minipage}\hfill
\begin{minipage}[t]{0.20\linewidth}
\vspace{-3.0cm}
\vtt{satisfied 3/3}\\
\vtt{gratuitous 1}\\
\vtt{executes yes}\\[2pt]
\textbf{Not exact.}
\end{minipage}

\smallskip
The fine-tuned model makes the requested change and also rewrites the value at 1.0:
\begin{center}
\vtt{mrr = [0.25, 0.259, 0.257, 0.255, 0.256, 0.258]}\\
\vtt{$\rightarrow$ [\textbf{0.256}, 0.259, 0.257, 0.255, \textbf{0.260}, 0.258]}
\end{center}
The value at 1.8 is the request, and the value at 1.0 is outside it, so the judge records one
gratuitous change. The base model changes only the value at 1.8 and is an exact edit.
\end{tcolorbox}

\begin{tcolorbox}[failcard={Partial completion: one requirement met, one left undone},breakable=false]
\textbf{ChartEdit task 585.} \emph{``Increase the width of the actual line to 6 and change its
color to green to make the distinction between the actual and ideal lines more pronounced.''}
The instruction specifies two requirements: width 6 and color green.

\smallskip
\begin{minipage}[t]{0.38\linewidth}
\includegraphics[width=\linewidth]{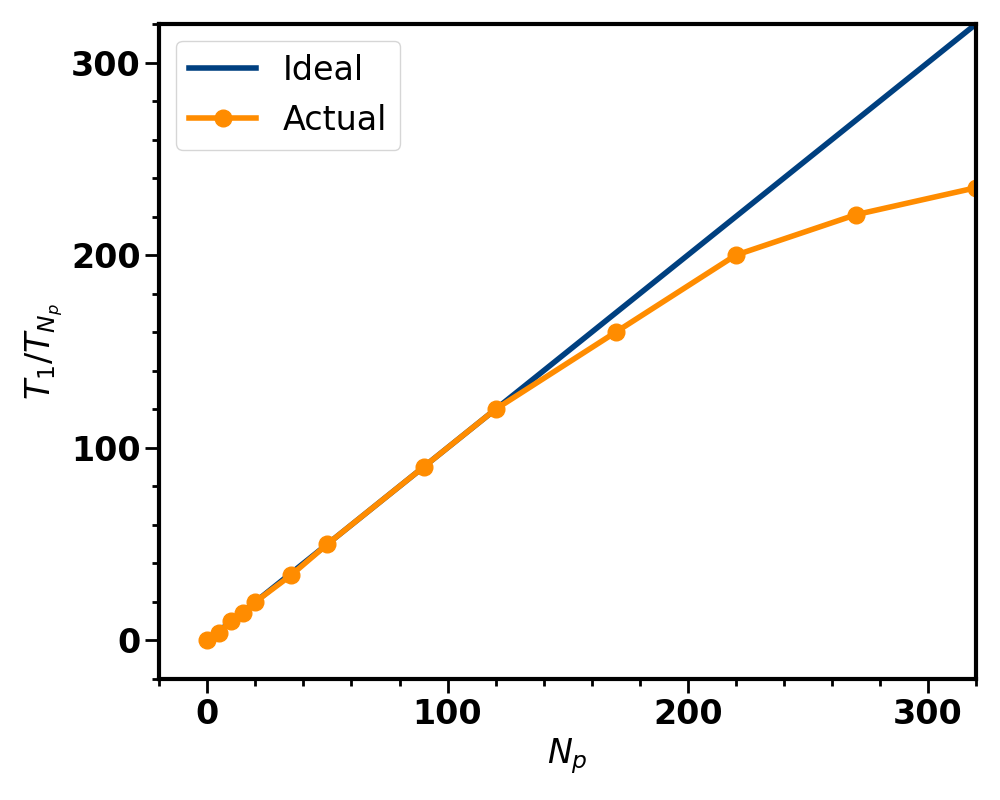}
\\{\footnotesize source}
\end{minipage}\hfill
\begin{minipage}[t]{0.38\linewidth}
\includegraphics[width=\linewidth]{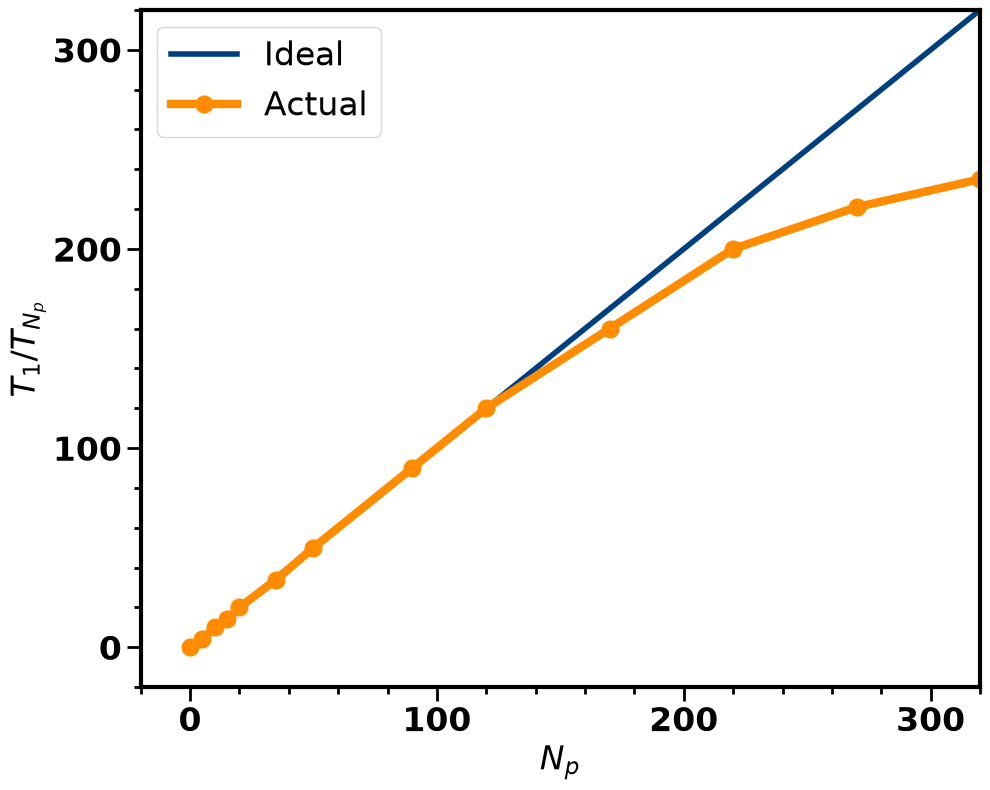}
\\{\footnotesize base model output}
\end{minipage}\hfill
\begin{minipage}[t]{0.20\linewidth}
\vspace{-3.0cm}
\vtt{satisfied 1/2}\\
\vtt{gratuitous 0}\\
\vtt{executes yes}\\[2pt]
\textbf{Not exact.}
\end{minipage}

\smallskip
The base model changes \vtt{linewidth=4} to \vtt{linewidth=6} but keeps \vtt{color='\#FF8C00'},
so the line is thicker and still orange. The judge marks the width requirement satisfied and
the color requirement unmet. The fine-tuned model makes both changes and is an exact edit.
\end{tcolorbox}

\begin{tcolorbox}[conceptcard={Coupled-update: a recolor whose legend must follow},breakable=false]
\textbf{ChartEdit task 451.} \emph{``Change the color of the kNN-prompting bars from royalblue to
orange for better differentiation.''}

\smallskip
The bars and the hand-built legend use separate color settings, so the legend does
not update automatically:
\begin{center}
\vtt{colors = ['red', 'green', \textbf{'royalblue'}, 'purple']}\\
\vtt{legend\_elements = [..., Patch(color=\textbf{'blue'}, label='kNN-prompting'), ...]}
\end{center}

\begin{center}
\includegraphics[width=0.9\linewidth]{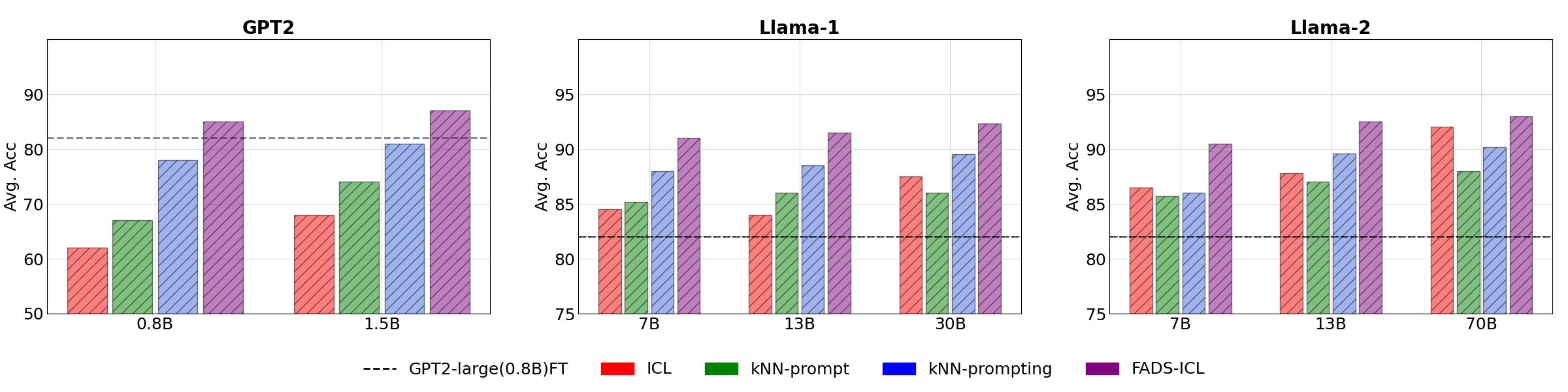}
\\{\footnotesize Source: blue bars and a blue legend patch.}
\end{center}

\textbf{R1 (explicit).} The kNN-prompting bars are orange.\\
\textbf{C1 (coupled).} The legend patch labelled kNN-prompting is orange.

\smallskip
Changing \vtt{colors[2]} implements R1. C1 requires updating the separate
\vtt{legend\_elements[3]} patch, although the instruction does not mention the legend.

\begin{center}
\includegraphics[width=0.9\linewidth]{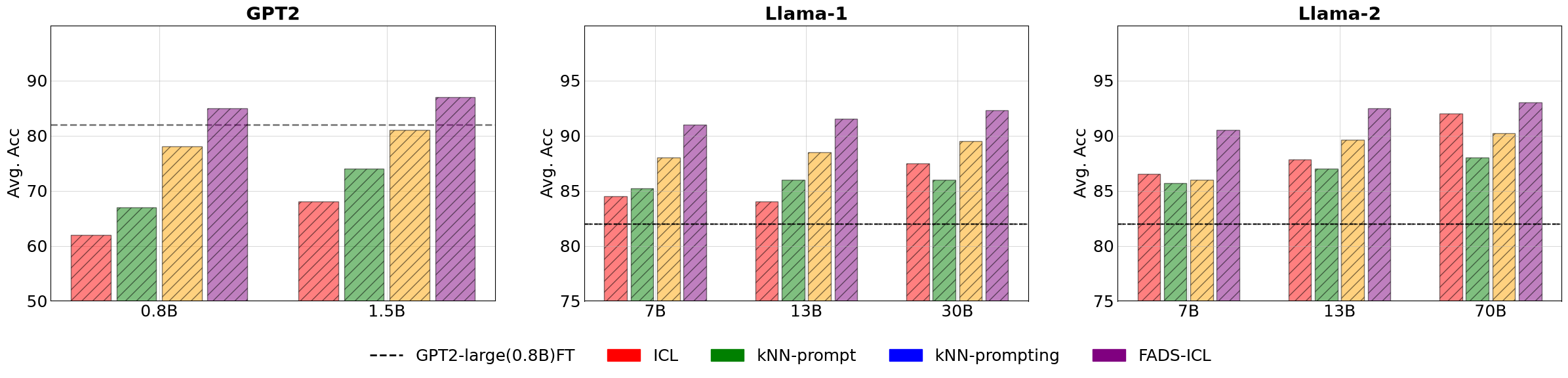}
\\{\footnotesize \textbf{Missed coupled update (constructed illustration):}
orange bars, blue legend; R1 is met, C1 is unmet.}
\end{center}

\begin{center}
\includegraphics[width=0.9\linewidth]{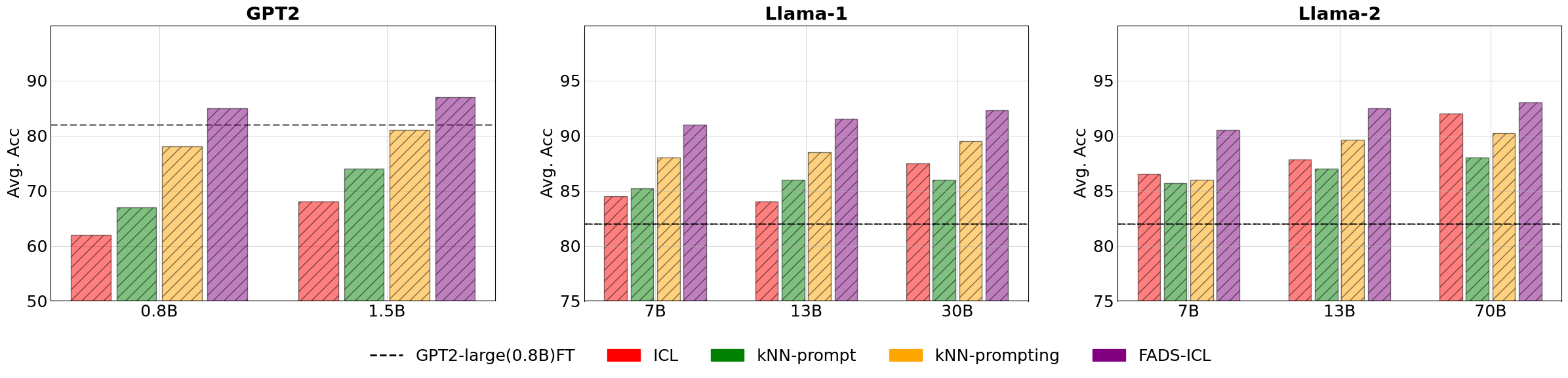}
\\{\footnotesize \textbf{Completed coupled update (illustration):}
orange bars and legend; both R1 and C1 are met.}
\end{center}
\end{tcolorbox}
\FloatBarrier
\section{Supplementary Evaluation Results}
\label{app:eval-supplement}

We present additional results for benchmark-facing metrics and our evaluation protocol.

\subsection{Complete benchmark-facing transfer results}
\label{app:ood-benchmark-facing}

\Cref{tab:ood-full-results} extends \Cref{tab:benchmark-results} with ChartScore from ChartEdit and the
native 8D F1, LLM-score, and LMM-score columns from Chart2Code. 
CodeScore improves in nearly every comparison, while ChartScore is mixed and ExecRate is the
least stable. CodeScore also scores programs that do not execute, and the LLM and LMM columns use
benchmark-specific scoring subsets, so none of these columns is a task-level success rate.
\Cref{tab:holistic-outcomes} compares holistic edit verdicts with requirement completion.
CR$_\mu$ is the pooled requirement recall defined in \Cref{sec:exact-edit-protocol};
CR$_M=\frac{1}{N}\sum_{i=1}^{N}\mathrm{CR}_i$ is the mean per-task completion ratio over
$N$ tasks. Both completion measures improve in 19 of 20 comparisons after fine-tuning.

\begin{table*}[t]
\centering
\scriptsize
\setlength{\tabcolsep}{2.5pt}
\renewcommand{\arraystretch}{0.88}
\caption{Complete out-of-domain benchmark-facing transfer results.}
\label{tab:ood-full-results}
\resizebox{\linewidth}{!}{
\begin{tabular}{llrrrrrr}
\toprule
Benchmark & Family & ExecRate & CodeScore & ChartScore & 8D F1 & LLM score & LMM score \\
\midrule
\multirow{5}{*}{\textbf{ChartEdit}}
 & Phi-3.5-V & $0.860\to0.860$ & $63.8\to\mathbf{75.4}$ & $91.6\to\mathbf{92.4}$ & $91.60\to\mathbf{92.54}$ & $90.91\to\mathbf{91.69}$ & $44.88\to\mathbf{54.80}$ \\
 & InternVL2.5 & $0.900\to0.860$ & $51.2\to\mathbf{71.6}$ & $90.2\to\mathbf{92.0}$ & $90.03\to\mathbf{92.07}$ & $90.53\to90.29$ & $41.82\to\mathbf{51.60}$ \\
 & Qwen3.5-4B & $0.900\to\mathbf{0.910}$ & $83.1\to\mathbf{86.7}$ & $94.0\to\mathbf{94.5}$ & $93.69\to\mathbf{94.13}$ & $94.21\to93.79$ & $59.16\to\mathbf{64.36}$ \\
 & Qwen3.5-9B & $0.940\to0.920$ & $83.8\to\mathbf{89.8}$ & $94.7\to\mathbf{95.1}$ & $93.81\to\mathbf{94.40}$ & $94.72\to94.32$ & $65.24\to\mathbf{65.98}$ \\
 & Granite & $0.870\to0.860$ & $68.5\to\mathbf{79.7}$ & $90.7\to\mathbf{93.6}$ & $90.26\to\mathbf{93.17}$ & $90.09\to\mathbf{92.29}$ & $49.72\to\mathbf{58.16}$ \\
\cmidrule(lr){1-8}
\multirow{5}{*}{\textbf{Chart2Code-L2}}
 & Phi-3.5-V & $0.400\to0.334$ & $27.6\to\mathbf{38.0}$ & $54.2\to\mathbf{58.9}$ & $23.99\to21.17$ & $63.77\to\mathbf{67.18}$ & $3.31\to\mathbf{3.41}$ \\
 & InternVL2.5 & $0.594\to0.380$ & $20.7\to\mathbf{32.6}$ & $53.7\to52.0$ & $31.46\to22.17$ & $61.84\to\mathbf{65.72}$ & $2.74\to\mathbf{3.12}$ \\
 & Qwen3.5-4B & $0.414\to\mathbf{0.444}$ & $60.5\to\mathbf{62.1}$ & $67.6\to\mathbf{69.0}$ & $28.76\to\mathbf{31.09}$ & $77.60\to\mathbf{78.33}$ & $7.50\to\mathbf{7.67}$ \\
 & Qwen3.5-9B & $0.567\to0.540$ & $66.7\to\mathbf{73.4}$ & $69.5\to\mathbf{72.1}$ & $39.50\to\mathbf{39.80}$ & $78.32\to\mathbf{79.70}$ & $7.92\to\mathbf{9.01}$ \\
 & Granite & $0.596\to0.373$ & $29.7\to\mathbf{43.1}$ & $56.0\to\mathbf{59.1}$ & $33.78\to23.91$ & $65.23\to\mathbf{70.78}$ & $3.09\to\mathbf{4.03}$ \\
\cmidrule(lr){1-8}
\multirow{5}{*}{\textbf{ChartM$^3$}}
 & Phi-3.5-V & $0.660\to0.650$ & $35.1\to\mathbf{46.3}$ & $89.9\to\mathbf{90.3}$ & $92.74\to92.23$ & $91.36\to\mathbf{92.56}$ & $35.45\to\mathbf{43.43}$ \\
 & InternVL2.5 & $0.690\to\mathbf{0.740}$ & $32.7\to\mathbf{43.3}$ & $91.3\to89.6$ & $92.50\to92.25$ & $92.32\to\mathbf{93.40}$ & $39.31\to\mathbf{42.67}$ \\
 & Qwen3.5-4B & $0.660\to\mathbf{0.700}$ & $53.7\to\mathbf{66.8}$ & $92.4\to91.8$ & $94.54\to93.09$ & $94.49\to\mathbf{94.98}$ & $46.91\to\mathbf{54.61}$ \\
 & Qwen3.5-9B & $0.730\to0.720$ & $61.2\to\mathbf{74.8}$ & $94.3\to93.1$ & $94.82\to93.95$ & $95.02\to\mathbf{95.85}$ & $56.03\to\mathbf{57.44}$ \\
 & Granite & $0.700\to0.630$ & $43.9\to\mathbf{48.4}$ & $90.3\to86.1$ & $91.56\to90.19$ & $91.18\to\mathbf{93.32}$ & $40.33\to\mathbf{45.17}$ \\
\cmidrule(lr){1-8}
\multirow{5}{*}{\textbf{ChartSync}}
 & Phi-3.5-V & $0.940\to0.930$ & $84.1\to\mathbf{85.5}$ & $97.9\to\mathbf{98.0}$ & $97.18\to96.76$ & $97.77\to\mathbf{98.14}$ & $77.57\to\mathbf{81.52}$ \\
 & InternVL2.5 & $0.960\to0.930$ & $79.8\to\mathbf{88.2}$ & $98.3\to96.9$ & $97.29\to96.11$ & $98.13\to\mathbf{98.22}$ & $74.08\to\mathbf{86.06}$ \\
 & Qwen3.5-4B & $0.990\to0.980$ & $92.3\to\mathbf{92.6}$ & $98.9\to98.8$ & $97.82\to97.61$ & $99.24\to97.33$ & $87.59\to87.42$ \\
 & Qwen3.5-9B & $0.990\to0.980$ & $95.7\to94.8$ & $99.5\to99.3$ & $97.85\to97.81$ & $99.45\to99.26$ & $92.17\to91.06$ \\
 & Granite & $0.870\to\mathbf{0.900}$ & $79.7\to\mathbf{83.4}$ & $97.6\to97.3$ & $96.50\to\mathbf{96.59}$ & $92.00\to\mathbf{97.66}$ & $75.33\to\mathbf{81.13}$ \\
\bottomrule
\end{tabular}}
\end{table*}

\begin{table}[t]
\centering
\scriptsize
\setlength{\tabcolsep}{4pt}
\caption{Complementary holistic outcomes.}
\label{tab:holistic-outcomes}
\begin{tabular}{llrrrr}
\toprule
Benchmark & Family & AppliedRate $\uparrow$ & EditFidelity $\uparrow$ & CR$_\mu$ $\uparrow$ & CR$_M$ $\uparrow$ \\
\midrule
\multirow{5}{*}{\textbf{ChartEdit}}
 & Phi-3.5-V & \textbf{.625}\,\gain{+.127} & \textbf{.389}\,\gain{+.121} & \textbf{.773}\,\gain{+.138} & \textbf{.758}\,\gain{+.145} \\
 & InternVL2.5 & \textbf{.573}\,\gain{+.224} & \textbf{.345}\,\gain{+.143} & \textbf{.707}\,\gain{+.234} & \textbf{.703}\,\gain{+.241} \\
 & Qwen3.5-4B & \textbf{.711}\,\gain{+.024} & \textbf{.470}\,\gain{+.052} & \textbf{.867}\,\gain{+.047} & \textbf{.871}\,\gain{+.049} \\
 & Qwen3.5-9B & \textbf{.743}\,\gain{+.059} & \textbf{.498}\,\gain{+.022} & \textbf{.895}\,\gain{+.068} & \textbf{.899}\,\gain{+.073} \\
 & Granite & \textbf{.666}\,\gain{+.131} & \textbf{.384}\,\gain{+.110} & \textbf{.812}\,\gain{+.114} & \textbf{.811}\,\gain{+.117} \\
\cmidrule(lr){1-6}
\multirow{5}{*}{\textbf{Chart2Code-L2}}
 & Phi-3.5-V & \textbf{.234}\,\gain{+.136} & \textbf{.032}\,\gain{+.015} & \textbf{{.691}}\,\gain{+.120} & \textbf{.699}\,\gain{+.156} \\
 & InternVL2.5 & \textbf{.160}\,\gain{+.098} & \textbf{.024}\,\gain{+.006} & \textbf{{.624}}\,\gain{{+.199}} & \textbf{.630}\,\gain{+.196} \\
 & Qwen3.5-4B & \textbf{.469}\,\gain{+.054} & \textbf{.131}\,\gain{+.029} & \textbf{{.844}}\,\gain{{+.013}} & \textbf{.847}\,\gain{+.004} \\
 & Qwen3.5-9B & \textbf{.506}\,\gain{+.104} & \textbf{.170}\,\gain{+.037} & \textbf{{.896}}\,\gain{{+.053}} & \textbf{.894}\,\gain{+.048} \\
 & Granite & \textbf{.263}\,\gain{+.170} & \textbf{.058}\,\gain{+.030} & \textbf{{.723}}\,\gain{{+.206}} & \textbf{.725}\,\gain{+.199} \\
\cmidrule(lr){1-6}
\multirow{5}{*}{\textbf{ChartM$^3$}}
 & Phi-3.5-V & \textbf{.243}\,\gain{+.047} & \textbf{.105}\,\gain{+.031} & \textbf{.390}\,\gain{+.097} & \textbf{.427}\,\gain{+.120} \\
 & InternVL2.5 & \textbf{.214}\,\gain{+.086} & \textbf{.117}\,\gain{+.050} & \textbf{.338}\,\gain{+.088} & \textbf{.363}\,\gain{+.116} \\
 & Qwen3.5-4B & \textbf{.343}\,\gain{+.092} & \textbf{.175}\,\gain{+.050} & \textbf{.623}\,\gain{+.150} & \textbf{.662}\,\gain{+.152} \\
 & Qwen3.5-9B & \textbf{.349}\,\gain{+.097} & \textbf{.186}\,\gain{+.044} & \textbf{.701}\,\gain{+.181} & \textbf{.724}\,\gain{+.167} \\
 & Granite & \textbf{.240}\,\gain{+.011} & \textbf{.111}\,\gain{+.012} & \textbf{.409}\,\gain{+.041} & \textbf{.457}\,\gain{+.068} \\
\cmidrule(lr){1-6}
\multirow{5}{*}{\textbf{ChartSync}}
 & Phi-3.5-V & \textbf{.835}\,\gain{+.058} & \textbf{.690}\,\gain{+.033} & \textbf{.819}\,\gain{+.056} & \textbf{.847}\,\gain{+.043} \\
 & InternVL2.5 & \textbf{.872}\,\gain{+.161} & \textbf{.747}\,\gain{+.133} & \textbf{.838}\,\gain{+.119} & \textbf{.867}\,\gain{+.145} \\
 & Qwen3.5-4B & \textbf{.874}\,\gain{+.010} & \textbf{.807}\,\gain{+.008} & .886\,\loss{-.002} & \textbf{.891}\,\gain{+.004} \\
 & Qwen3.5-9B & .911\,\loss{-.010} & .840\,\loss{-.021} & \textbf{.927}\,\gain{+.011} & .932\,\tie{+.000} \\
 & Granite & \textbf{.828}\,\gain{+.024} & \textbf{.680}\,\gain{+.093} & \textbf{.819}\,\gain{+.036} & \textbf{.825}\,\gain{+.019} \\
\midrule
\multicolumn{2}{l}{\textit{improved in}} & \textit{19/20} & \textit{19/20} & \textit{19/20} & \textit{19/20} \\
\bottomrule
\end{tabular}
\end{table}

\subsection{Protocol diagnostics}
\label{app:exec-error-audit}

\begin{figure}[!t]
  \centering
  \includegraphics[width=0.78\linewidth]{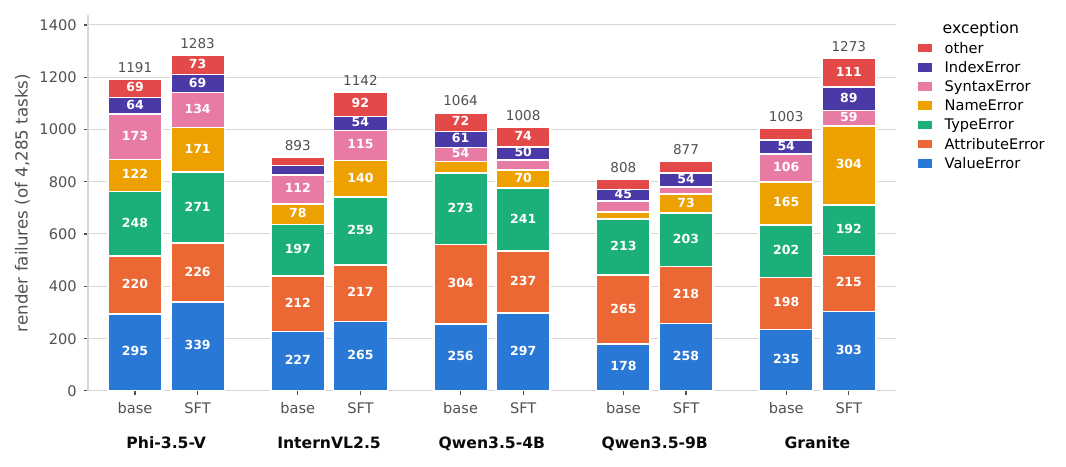}
  \vspace{-.3cm}
  \caption{Terminal Python exception counts in the 4{,}285-task diagnostic run for all
  five models.}
  \label{fig:error-stack}
  \vspace{-.5cm}
\end{figure}

\Cref{tab:chartsync-native-full} reports ChartSync's native text, geometry, and whole-output
scores as complementary diagnostics. \Cref{fig:error-stack} summarizes terminal Python
exception counts in the diagnostic run. These coarse exception labels describe execution
failures.

\begin{table*}[h]
\centering
\small
\setlength{\tabcolsep}{3.2pt}
\renewcommand{\arraystretch}{1.08}
\caption{ChartSync diagnostic metrics.}
\label{tab:chartsync-native-full}
\resizebox{\linewidth}{!}{
\begin{tabular}{p{2.25cm}p{2.55cm}p{2.55cm}p{2.55cm}p{2.55cm}p{2.55cm}}
\toprule
Metric & Phi-3.5-V & InternVL2.5 & Qwen3.5-4B & Qwen3.5-9B & Granite \\
\midrule
TESR & $81.2\to\mathbf{88.0}$ & $74.3\to\mathbf{90.7}$ & $89.4\to\mathbf{91.7}$ & $94.7\to94.1$ & $83.7\to\mathbf{87.9}$ \\
VLCS & $64.8\to\mathbf{78.5}$ & $76.5\to\mathbf{84.6}$ & $90.0\to87.6$ & $87.3\to\mathbf{92.3}$ & $79.0\to\mathbf{84.9}$ \\
BFS & $95.4\to93.6$ & $96.2\to94.7$ & $98.1\to97.4$ & $98.5\to98.3$ & $91.1\to\mathbf{94.3}$ \\
OCR F1 & $93.3\to92.8$ & $94.0\to92.5$ & $94.1\to93.9$ & $94.6\to94.3$ & $93.7\to92.9$ \\
SSIM & $88.4\to87.8$ & $88.4\to88.3$ & $88.8\to88.8$ & $88.9\to\mathbf{89.0}$ & $88.2\to88.2$ \\
\bottomrule
\end{tabular}}
\end{table*}

\subsection{Reference-program calibration}
\label{app:reference-calibration}

\Cref{tab:ceilings} reports how released benchmark references score under our renderer and
judge pipeline.
{For ChartM$^3$, the references obtain an AppliedRate of 0.345 despite an execution
rate of 0.91 and a ChartScore of 89.26. AppliedRate checks whether the requested edit
reaches its named target, whereas ChartScore assesses overall rendered-chart quality.
These aggregate scores alone do not distinguish reference-edit errors from errors
introduced by benchmark adaptation or automatic judging.}

\begin{table}[t]
\centering
\small
\caption{Reference-program calibration. These calibrate the renderer--judge pipeline and
are not model-specific capability ceilings.}
\label{tab:ceilings}
\resizebox{\linewidth}{!}{
\begin{tabular}{lrrrrr}
\toprule
Benchmark & ExecRate & CodeScore & ChartScore & AppliedRate & EditFidelity \\
\midrule
ChartEdit & 1.00 & 95.41 & 99.88 & 0.796 & 0.552 \\
Chart2Code L2 & 0.97 & 98.04 & 95.92 & 0.757 & 0.455 \\
ChartSync & 0.99 & 99.79 & 98.58 & 0.956 & 0.901 \\
ChartM$^3$ & 0.91 & 89.52 & 89.26 & 0.345 & 0.297 \\
\bottomrule
\end{tabular}}
\end{table}

\FloatBarrier
\subsection{Human validation of exact-edit judgments}
\label{app:human-transfer}

The results in \Cref{tab:evaluator-audit} cover 200 paired external tasks: 69 from ChartEdit,
50 from ChartM$^3$, 64 from ChartSync, and 17 from Chart2Code-L2. Each task pairs a base
output with its fine-tuned counterpart. Human judgments are made on blinded outputs
using the labels described below.
Agreement and gains in \Cref{tab:evaluator-audit} use exact-edit verdicts. A positive
human exact-edit verdict requires every listed atomic requirement to be satisfied and no
gratuitous change to unrelated chart content. All outputs in this audit render successfully.

\begin{table}[!h]
\centering
\small
\setlength{\tabcolsep}{4pt}
\caption{Paired human--protocol exact-edit agreement and fine-tuning gains on rendered outputs.
Agreement is measured per output; $\Delta_P$ and $\Delta_H$ are the gains under automatic
and human judgments, respectively. Gains are absolute rate differences.}
\label{tab:evaluator-audit}
\begin{tabular}{lrrrr}
\toprule
Benchmark & Pairs  & Agreement & $\Delta_P$ & $\Delta_H$ \\
\midrule
ChartEdit     & 69   & 91\% & $+0.232$ & $+0.203$ \\
ChartM$^3$    & 50  & 91\% & $+0.180$ & $+0.160$ \\
ChartSync     & 64   & 98\% & $+0.031$ & $+0.078$ \\
Chart2Code-L2 & 17  & 91\% & $+0.059$ & $+0.118$ \\
\cmidrule(lr){1-5}
\textit{All}  & 200  & 93\% & $+0.140$ & $+0.145$ \\
\bottomrule
\end{tabular}
\end{table}

\paragraph{Audit procedure.}
\begin{itemize}
  \item \textbf{Sampling.} {The pairs come from a blinded annotation package of
  370 tasks drawn with a fixed seed from two model families, InternVL2.5 and
  Qwen3.5-4B, chosen as the arms with the largest and smallest automatic gains. Sampling is
  stratified by benchmark, and Chart2Code-L2 is further stratified into four bins of
  explicit-requirement count and oversampled, with per-item weights that restore population
  proportions. The 200 audited pairs are those in which both outputs render: all 96 such pairs
  from the first sampling round and 104 drawn at random. By family, ChartEdit contributes 33 InternVL2.5 and 36 Qwen3.5-4B pairs,
  ChartM$^3$ 30 and 20, ChartSync 31 and 33, and Chart2Code-L2 10 and 7. Two tasks appear
  under both families, and no task appears twice within a family.}
  \item \textbf{Annotators.} Two computer science students with
  experience using Python plotting libraries conduct the annotation. Their labels are
  saved before any comparison.
  \item \textbf{Evidence and blinding.} Annotators see the instruction, the
  source program and its rendering, the fixed atomic-requirement list used by the automatic
  protocol, and two edited outputs, labelled A and B, each with its program and rendering. Which output is the
  fine-tuned one is randomized independently per item, and the mapping is kept off the
  annotation server. Model families appear only as pseudonyms, and judge verdicts and
  rationales are stored for later comparison but never shown to annotators.
  \item \textbf{Labels.} Annotators judge A and B separately against the same requirement
  list, labelling each requirement satisfied, violated, or ambiguous, and record whether the
  output makes any gratuitous change to unrelated content in the source chart. Necessary
  consequences of a requested change do not count as gratuitous. The human
  \emph{exact-edit} verdict is positive only when every listed requirement is satisfied
  and no gratuitous change is present. The audit records requirement-level labels and
  a task-level gratuitous-change flag.
\end{itemize}

\FloatBarrier
\subsection{Effect of exactness filtering}
\label{app:filter-ablation}

Motivated by~\citep{chen2024alpagasus}, which explored the effects of data quantity and quality on training, \Cref{tab:filtering} compares fine-tuning on the corpus before and after exactness
filtering on all four external benchmarks. On ChartM$^3$, coupled-update recall improves by 3.0 and
6.9 percentage points for Phi-3.5-V and Qwen3.5-4B, respectively. On Chart2Code-L2,
Qwen3.5-4B improves requirement recall and full completion by 2.5 and 3.3 percentage
points, while exact-edit rate remains stable.

\begin{table}[!ht]
\centering
\small
\setlength{\tabcolsep}{3.5pt}
\caption{Fine-tuning on the corpus before and after exactness filtering across all four
external benchmarks.}
\label{tab:filtering}
\resizebox{\linewidth}{!}{%
\begin{tabular}{lllrrrrr}
\toprule
Benchmark & Model & Training data & \shortstack{Requirement\\recall} $\uparrow$ & \shortstack{Full\\completion} $\uparrow$ & \shortstack{Gratuitous\\per task} $\downarrow$ & \shortstack{Coupled-update\\recall} $\uparrow$ & \shortstack{Exact-edit\\rate} $\uparrow$ \\
\midrule
\multirow{4}{*}{ChartEdit} & \multirow{2}{*}{Phi-3.5-V} & Before filtering & .764 & .670 & .348 & .527 & .418 \\
 &  & Released & \textbf{.773} & \textbf{.674} & \textbf{.346} & \textbf{.540} & \textbf{.425} \\
\cmidrule(lr){2-8}
 & \multirow{2}{*}{Qwen3.5-4B} & Before filtering & \textbf{.873} & .804 & \textbf{.140} & .666 & .566 \\
 &  & Released & .867 & .804 & .165 & \textbf{.688} & \textbf{.572} \\
\midrule
\multirow{4}{*}{Chart2Code-L2} & \multirow{2}{*}{Phi-3.5-V} & Before filtering & .689 & \textbf{.034} & \textbf{1.404} & .489 & .001 \\
 &  & Released & \textbf{.691} & .032 & 1.433 & \textbf{.493} & \textbf{.002} \\
\cmidrule(lr){2-8}
 & \multirow{2}{*}{Qwen3.5-4B} & Before filtering & .819 & .138 & .924 & .699 & \textbf{.045} \\
 &  & Released & \textbf{.844} & \textbf{.171} & \textbf{.900} & \textbf{.714} & .044 \\
\midrule
\multirow{4}{*}{ChartM$^3$} & \multirow{2}{*}{Phi-3.5-V} & Before filtering & .382 & \textbf{.325} & .941 & .318 & .142 \\
 &  & Released & \textbf{.390} & .321 & \textbf{.926} & \textbf{.348} & \textbf{.152} \\
\cmidrule(lr){2-8}
 & \multirow{2}{*}{Qwen3.5-4B} & Before filtering & .620 & .511 & \textbf{.401} & .484 & .304 \\
 &  & Released & \textbf{.623} & \textbf{.540} & .407 & \textbf{.553} & \textbf{.331} \\
\midrule
\multirow{4}{*}{ChartSync} & \multirow{2}{*}{Phi-3.5-V} & Before filtering & \textbf{.837} & \textbf{.841} & \textbf{.183} & .809 & \textbf{.679} \\
 &  & Released & .819 & .832 & .210 & \textbf{.815} & .668 \\
\cmidrule(lr){2-8}
 & \multirow{2}{*}{Qwen3.5-4B} & Before filtering & \textbf{.894} & \textbf{.885} & .082 & \textbf{.839} & .793 \\
 &  & Released & .886 & .879 & \textbf{.062} & .825 & \textbf{.800} \\
\bottomrule
\end{tabular}}
\end{table}

\FloatBarrier
\section{Dataset Release and Usage}
\label{app:release}

This section documents the collection-ledger fields and datasheet notes.

\subsection{Record schema}
\label{app:schema}

The collection ledger stores source and edited programs, available renderings, and
\texttt{evaluation.json}. Identifiers such as \texttt{task\_id}, \texttt{pool\_idx},
\texttt{source\_chart\_id}, \texttt{tmpl\_id}, and \texttt{binding\_id} link a candidate to its
source and template; category, target object, chart type, library, and instruction describe
the editing task. Timing and \texttt{attempt\_log} retain generation, execution, validation,
and repair history. The \texttt{preserved} field stores the first-pass judge's
\texttt{no\_additional\_change} decision.

The mechanical signals \texttt{patch\_nontrivial}, \texttt{api\_touch},
\texttt{render\_changed}, and \texttt{pixel\_dist} provide diagnostic evidence. In particular, a nontrivial code diff does not establish that the
requested chart property changed. First-pass flags and later requirement-level audits serve
different roles; a positive first-pass flag alone does not certify final exactness.

\subsection{Datasheet notes}
\label{app:datasheet}

\paragraph{Collection.} Source programs are drawn from ChartNet~\citep{kondic2026chartnet};
instructions and edited programs are generated by our pipeline. Coverage is non-uniform
across the taxonomy.

\paragraph{Models.} Editing and first-pass judging use \texttt{gemma-4-31B-it} served locally
(vLLM, W4A16 quantized); the cross-audit uses \texttt{Qwen3.8-27B} with the same exact-edit protocol
(\Cref{app:exact-prompts}). 

\paragraph{Compute.} Per-sample cost is dominated by editor generation (mean 66\,s) and
judging (mean 17\,s); rendering is under 1\,s. The full collection is on the order of
$1.8\times10^5$ editor calls and an equal number of judge calls.

%% file: iclr2027_conference.bib
@inproceedings{zhao2025chartedit,
  title={Chartedit: How far are mllms from automating chart analysis? evaluating mllms’ capability via chart editing},
  author={Zhao, Xuanle and Liu, Xuexin and Haoyue, Yang and Luo, Xianzhen and Zeng, Fanhu and Li, Jianling and Shi, Qi and Chen, Chi},
  booktitle={Findings of the Association for Computational Linguistics: ACL 2025},
  pages={3616--3630},
  year={2025}
}

@inproceedings{yang2025chartm3,
  title={Chartm3: Benchmarking chart editing with multimodal instructions},
  author={Yang, Donglu and Zhang, Liang and Yue, Zihao and Chen, Liangyu and Xu, Yichen and Wang, Wenxuan and Jin, Qin},
  booktitle={Proceedings of the 33rd ACM International Conference on Multimedia},
  pages={5001--5009},
  year={2025}
}

@inproceedings{chen2026charteditor,
  title={Charteditor: A reinforcement learning framework for robust chart editing},
  author={Chen, Liangyu and Xu, Yichen and Ma, Jianzhe and Liu, Yuqi and Yang, Donglu and Zhang, Liang and Yue, Zihao and Wang, Wenxuan and Jin, Qin},
  booktitle={Proceedings of the AAAI Conference on Artificial Intelligence},
  pages={20199--20207},
  year={2026}
}

@article{kapadnis2026charteditbench,
  title={ChartEditBench: Evaluating Grounded Multi-Turn Chart Editing in Multimodal Language Models},
  author={Kapadnis, Manav Nitin and Baghel, Lawanya and Naik, Atharva and Ros{\'e}, Carolyn},
  journal={arXiv preprint arXiv:2602.15758},
  year={2026}
}

@inproceedings{li2026charts,
  title={Charts are not images: On the challenges of scientific chart editing},
  author={Li, Li and Rossi, Ryan and Kim, Sungchul and Choudhary, Sunav and Dernoncourt, Franck and Mathur, Puneet and Tu, Zhengzhong and Zhao, Yue},
  booktitle={International Conference on Learning Representations},
  volume={2026},
  pages={11937--11966},
  year={2026}
}

@inproceedings{zhao2025chartcoder,
  title={Chartcoder: Advancing multimodal large language model for chart-to-code generation},
  author={Zhao, Xuanle and Luo, Xianzhen and Shi, Qi and Chen, Chi and Wang, Shuo and Liu, Zhiyuan and Sun, Maosong},
  booktitle={Proceedings of the 63rd Annual Meeting of the Association for Computational Linguistics (Volume 1: Long Papers)},
  pages={7333--7348},
  year={2025}
}

@inproceedings{yang2025chartmimic,
  title={Chartmimic: Evaluating lmm's cross-modal reasoning capability via chart-to-code generation},
  author={Yang, Cheng and Shi, Chufan and Liu, Yaxin and Shui, Bo and Wang, Junjie and Jing, Mohan and Xu, Linran and Zhu, Xinyu and Li, Siheng and Zhang, Yuxiang and others},
  booktitle={International Conference on Learning Representations},
  volume={2025},
  pages={26590--26646},
  year={2025}
}

@article{he2026chart,
  title={Chart Specification: Structural Representations for Incentivizing VLM Reasoning in Chart-to-Code Generation},
  author={He, Minggui and Dai, Mingchen and Zhang, Jian and Liu, Yilun and Tao, Shimin and Zeng, Pufan and Yoshie, Osamu and Ieiri, Yuya},
  journal={arXiv preprint arXiv:2602.10880},
  year={2026}
}

@inproceedings{liu2023matcha,
  title={Matcha: Enhancing visual language pretraining with math reasoning and chart derendering},
  author={Liu, Fangyu and Piccinno, Francesco and Krichene, Syrine and Pang, Chenxi and Lee, Kenton and Joshi, Mandar and Altun, Yasemin and Collier, Nigel and Eisenschlos, Julian},
  booktitle={Proceedings of the 61st Annual Meeting of the Association for Computational Linguistics (Volume 1: Long Papers)},
  pages={12756--12770},
  year={2023}
}

@inproceedings{masry2023unichart,
  title={Unichart: A universal vision-language pretrained model for chart comprehension and reasoning},
  author={Masry, Ahmed and Kavehzadeh, Parsa and Hoque, Enamul and Joty, Shafiq and others},
  booktitle={Proceedings of the 2023 conference on empirical methods in natural language processing},
  pages={14662--14684},
  year={2023}
}

@article{han2023chartllama,
  title={Chartllama: A multimodal llm for chart understanding and generation},
  author={Han, Yucheng and Zhang, Chi and Chen, Xin and Yang, Xu and Wang, Zhibin and Yu, Gang and Fu, Bin and Zhang, Hanwang},
  journal={arXiv preprint arXiv:2311.16483},
  year={2023}
}

@article{xia2025chartx,
  title={Chartx \& chartvlm: A versatile benchmark and foundation model for complicated chart reasoning},
  author={Xia, Renqiu and Ye, Hancheng and Yan, Xiangchao and Liu, Qi and Zhou, Hongbin and Chen, Zijun and Shi, Botian and Yan, Junchi and Zhang, Bo},
  journal={IEEE Transactions on Image Processing},
  year={2025},
  publisher={IEEE}
}

@inproceedings{masry2025chartgemma,
  title={Chartgemma: Visual instruction-tuning for chart reasoning in the wild},
  author={Masry, Ahmed and Thakkar, Megh and Bajaj, Aayush and Kartha, Aaryaman and Hoque, Enamul and Joty, Shafiq},
  booktitle={Proceedings of the 31st International Conference on Computational Linguistics: Industry Track},
  pages={625--643},
  year={2025}
}

@inproceedings{masry-etal-2022-chartqa,
    title = "{C}hart{QA}: A Benchmark for Question Answering about Charts with Visual and Logical Reasoning",
    author = "Masry, Ahmed  and
      Long, Do Xuan  and
      Tan, Jia Qing  and
      Joty, Shafiq  and
      Hoque, Enamul",
    editor = "Muresan, Smaranda  and
      Nakov, Preslav  and
      Villavicencio, Aline",
    booktitle = "Findings of the Association for Computational Linguistics: ACL 2022",
    month = may,
    year = "2022",
    address = "Dublin, Ireland",
    publisher = "Association for Computational Linguistics",
    url = "https://aclanthology.org/2022.findings-acl.177/",
    doi = "10.18653/v1/2022.findings-acl.177",
    pages = "2263--2279",
}

@article{zhang2023magicbrush,
  title={Magicbrush: A manually annotated dataset for instruction-guided image editing},
  author={Zhang, Kai and Mo, Lingbo and Chen, Wenhu and Sun, Huan and Su, Yu},
  journal={Advances in neural information processing systems},
  volume={36},
  pages={31428--31449},
  year={2023}
}

@inproceedings{sheynin2024emu,
  title={Emu edit: Precise image editing via recognition and generation tasks},
  author={Sheynin, Shelly and Polyak, Adam and Singer, Uriel and Kirstain, Yuval and Zohar, Amit and Ashual, Oron and Parikh, Devi and Taigman, Yaniv},
  booktitle={2024 ieee/cvf conference on computer vision and pattern recognition (cvpr)},
  pages={8871--8879},
  year={2024},
  organization={IEEE}
}

@incollection{wilkinson2011grammar,
  title={The grammar of graphics},
  author={Wilkinson, Leland},
  booktitle={Handbook of computational statistics: Concepts and methods},
  pages={375--414},
  year={2011},
  publisher={Springer}
}

@article{wickham2010layered,
  title={A layered grammar of graphics},
  author={Wickham, Hadley},
  journal={Journal of computational and graphical statistics},
  volume={19},
  number={1},
  pages={3--28},
  year={2010},
  publisher={Taylor \& Francis}
}

@article{satyanarayan2016vega,
  title={Vega-lite: A grammar of interactive graphics},
  author={Satyanarayan, Arvind and Moritz, Dominik and Wongsuphasawat, Kanit and Heer, Jeffrey},
  journal={IEEE transactions on visualization and computer graphics},
  volume={23},
  number={1},
  pages={341--350},
  year={2016},
  publisher={IEEE}
}

@inproceedings{tang2026charts,
  title={From charts to code: A hierarchical benchmark for multimodal models},
  author={Tang, Jiahao and Zhao, Henry Hengyuan and Wu, Lijian and Zhang, Zijian and Tao, Yifei and Mao, Dongxing and Wan, Yang and Tan, Jingru and Zeng, Min and Li, Min and others},
  booktitle={Proceedings of the 64th Annual Meeting of the Association for Computational Linguistics (Volume 1: Long Papers)},
  pages={13467--13566},
  year={2026}
}

@article{yu2026chartsync,
  title={ChartSync: A Benchmark for Visuo-Logical Cascading Chart Editing},
  author={Yu, Jiakang and Chai, Yixuan and Wang, Tianci and Jin, Rihui and Xu, Guangkai and Deng, Hongtao and Zhu, Xun and Gao, Wang and Guo, Xinrun and Wu, Haipang},
  journal={arXiv preprint arXiv:2607.10301},
  year={2026}
}

@article{kondic2026chartnet,
  title={Chartnet: A million-scale, high-quality multimodal dataset for robust chart understanding},
  author={Kondic, Jovana and Li, Pengyuan and Joshi, Dhiraj and Sanchez, Isaac and Wiesel, Ben and Abedin, Shafiq and Alfassy, Amit and Schwartz, Eli and Caraballo, Daniel and Cinar, Yagmur Gizem and others},
  journal={arXiv preprint arXiv:2603.27064},
  year={2026}
}

@inproceedings{yan2024chartreformer,
  title={Chartreformer: Natural language-driven chart image editing},
  author={Yan, Pengyu and Bhosale, Mahesh and Lal, Jay and Adhikari, Bikhyat and Doermann, David},
  booktitle={International Conference on Document Analysis and Recognition},
  pages={453--469},
  year={2024},
  organization={Springer}
}

@inproceedings{goswami2025plotedit,
  title={Plotedit: Natural language-driven accessible chart editing in pdfs via multimodal llm agents},
  author={Goswami, Kanika and Mathur, Puneet and Rossi, Ryan and Dernoncourt, Franck},
  booktitle={European Conference on Information Retrieval},
  pages={130--134},
  year={2025},
  organization={Springer}
}

@inproceedings{wu2025plot2code,
  title={Plot2code: A comprehensive benchmark for evaluating multi-modal large language models in code generation from scientific plots},
  author={Wu, Chengyue and Liang, Zhixuan and Ge, Yixiao and Guo, Qiushan and Lu, Zeyu and Wang, Jiahao and Shan, Ying and Luo, Ping},
  booktitle={Findings of the Association for Computational Linguistics: NAACL 2025},
  pages={3006--3028},
  year={2025}
}

@inproceedings{yang2024matplotagent,
  title={Matplotagent: Method and evaluation for llm-based agentic scientific data visualization},
  author={Yang, Zhiyu and Zhou, Zihan and Wang, Shuo and Cong, Xin and Han, Xu and Yan, Yukun and Liu, Zhenghao and Tan, Zhixing and Liu, Pengyuan and Yu, Dong and others},
  booktitle={Findings of the Association for Computational Linguistics: ACL 2024},
  pages={11789--11804},
  year={2024}
}

@article{satyanarayan2015reactive,
  title={Reactive vega: A streaming dataflow architecture for declarative interactive visualization},
  author={Satyanarayan, Arvind and Russell, Ryan and Hoffswell, Jane and Heer, Jeffrey},
  journal={IEEE transactions on visualization and computer graphics},
  volume={22},
  number={1},
  pages={659--668},
  year={2015},
  publisher={IEEE}
}

@inproceedings{brooks2023instructpix2pix,
  title={Instructpix2pix: Learning to follow image editing instructions},
  author={Brooks, Tim and Holynski, Aleksander and Efros, Alexei A},
  booktitle={2023 IEEE/CVF Conference on Computer Vision and Pattern Recognition (CVPR)},
  pages={18392--18402},
  year={2023},
  organization={IEEE}
}

@article{zheng2023judging,
  title={Judging llm-as-a-judge with mt-bench and chatbot arena},
  author={Zheng, Lianmin and Chiang, Wei-Lin and Sheng, Ying and Zhuang, Siyuan and Wu, Zhanghao and Zhuang, Yonghao and Lin, Zi and Li, Zhuohan and Li, Dacheng and Xing, Eric and others},
  journal={Advances in neural information processing systems},
  volume={36},
  pages={46595--46623},
  year={2023}
}

@inproceedings{chen2024internvl,
  title={Internvl: Scaling up vision foundation models and aligning for generic visual-linguistic tasks},
  author={Chen, Zhe and Wu, Jiannan and Wang, Wenhai and Su, Weijie and Chen, Guo and Xing, Sen and Zhong, Muyan and Zhang, Qinglong and Zhu, Xizhou and Lu, Lewei and others},
  booktitle={Proceedings of the IEEE/CVF Conference on Computer Vision and Pattern Recognition},
  pages={24185--24198},
  year={2024}
}

@misc{qwen3.5,
    title  = {{Qwen3.5}: Towards Native Multimodal Agents},
    author = {{Qwen Team}},
    month  = {February},
    year   = {2026},
    url    = {https://qwen.ai/blog?id=qwen3.5}
}

@misc{granite-vision-4.1-4b,
  title={Granite 4.1 Vision},
  author={{IBM Granite Vision Team}},
  year={2026},
  url={https://huggingface.co/ibm-granite/granite-vision-4.1-4b}
}

@misc{phi3,
      title={Phi-3 Technical Report: A Highly Capable Language Model Locally on Your Phone}, 
      author={Marah Abdin and others},
      year={2024},
      eprint={2404.14219},
      archivePrefix={arXiv},
      primaryClass={cs.CL},
      url={https://arxiv.org/abs/2404.14219}, 
}

@inproceedings{chen2024alpagasus,
  title={Alpagasus: Training a better alpaca with fewer data},
  author={Chen, Lichang and Li, Shiyang and Yan, Jun and Wang, Hai and Gunaratna, Kalpa and Yadav, Vikas and Tang, Zheng and Srinivasan, Vijay and Zhou, Tianyi and Huang, Heng and others},
  booktitle={International Conference on Learning Representations},
  volume={2024},
  pages={34767--34797},
  year={2024}
}

@inproceedings{liu2023g,
  title={G-eval: NLG evaluation using gpt-4 with better human alignment},
  author={Liu, Yang and Iter, Dan and Xu, Yichong and Wang, Shuohang and Xu, Ruochen and Zhu, Chenguang},
  booktitle={Proceedings of the 2023 conference on empirical methods in natural language processing},
  pages={2511--2522},
  year={2023}
}

@inproceedings{wang2024large,
  title={Large language models are not fair evaluators},
  author={Wang, Peiyi and Li, Lei and Chen, Liang and Cai, Zefan and Zhu, Dawei and Lin, Binghuai and Cao, Yunbo and Kong, Lingpeng and Liu, Qi and Liu, Tianyu and others},
  booktitle={Proceedings of the 62nd annual meeting of the association for computational linguistics (volume 1: Long papers)},
  pages={9440--9450},
  year={2024}
}

@article{chen2021evaluating,
  title={Evaluating large language models trained on code},
  author={Chen, Mark and Tworek, Jerry and Jun, Heewoo and Yuan, Qiming and Pinto, Henrique Ponde De Oliveira and Kaplan, Jared and Edwards, Harri and Burda, Yuri and Joseph, Nicholas and Brockman, Greg and others},
  journal={arXiv preprint arXiv:2107.03374},
  year={2021}
}

@article{panickssery2024llm,
  title={Llm evaluators recognize and favor their own generations},
  author={Panickssery, Arjun and Bowman, Samuel R and Feng, Shi},
  journal={Advances in Neural Information Processing Systems},
  volume={37},
  pages={68772--68802},
  year={2024}
}

@article{cassano2023can,
  title={Can it edit? evaluating the ability of large language models to follow code editing instructions},
  author={Cassano, Federico and Li, Luisa and Sethi, Akul and Shinn, Noah and Brennan-Jones, Abby and Ginesin, Jacob and Berman, Edward and Chakhnashvili, George and Lozhkov, Anton and Anderson, Carolyn Jane and others},
  journal={arXiv preprint arXiv:2312.12450},
  year={2023}
}

@inproceedings{jimenez2024swe,
  title={Swe-bench: Can language models resolve real-world github issues?},
  author={Jimenez, Carlos E and Yang, John and Wettig, Alexander and Yao, Shunyu and Pei, Kexin and Press, Ofir and Narasimhan, Karthik},
  booktitle={International Conference on Learning Representations},
  volume={2024},
  pages={54107--54157},
  year={2024}
}

@inproceedings{ku2024imagenhub,
  title={Imagenhub: Standardizing the evaluation of conditional image generation models},
  author={Ku, Max and Li, Tianle and Zhang, Kai and Lu, Yujie and Fu, Xingyu and Zhuang, Wenwen and Chen, Wenhu},
  booktitle={International Conference on Learning Representations},
  volume={2024},
  pages={46689--46722},
  year={2024}
}

@inproceedings{ku2024viescore,
  title={Viescore: Towards explainable metrics for conditional image synthesis evaluation},
  author={Ku, Max and Jiang, Dongfu and Wei, Cong and Yue, Xiang and Chen, Wenhu},
  booktitle={Proceedings of the 62nd Annual Meeting of the Association for Computational Linguistics (Volume 1: Long Papers)},
  pages={12268--12290},
  year={2024}
}

@inproceedings{cho2024davidsonian,
  title={Davidsonian scene graph: Improving reliability in fine-grained evaluation for text-to-image generation},
  author={Cho, Jaemin and Hu, Yushi and Baldridge, Jason and Garg, Roopal and Anderson, Peter and Krishna, Ranjay and Bansal, Mohit and Pont-Tuset, Jordi and Wang, Su},
  booktitle={International conference on learning representations},
  volume={2024},
  pages={15625--15645},
  year={2024}
}

@inproceedings{wang2023imagen,
  title={Imagen editor and editbench: Advancing and evaluating text-guided image inpainting},
  author={Wang, Su and Saharia, Chitwan and Montgomery, Ceslee and Pont-Tuset, Jordi and Noy, Shai and Pellegrini, Stefano and Onoe, Yasumasa and Laszlo, Sarah and Fleet, David J and Soricut, Radu and others},
  booktitle={2023 ieee/cvf conference on computer vision and pattern recognition (cvpr)},
  pages={18359--18369},
  year={2023},
  organization={IEEE}
}

@misc{qwen38,
    title = {{Qwen3.8-Max}: A New Bar for Coding and Cowork},
    url = {https://qwen.ai/blog?id=qwen3.8},
    author = {{Qwen Team}},
    month = {August},
    year = {2026}
}

@misc{gemmateam2026gemma4,
      title={Gemma 4 Technical Report}, 
      author={Gemma Team},
      year={2026},
      eprint={2607.02770},
      archivePrefix={arXiv},
      primaryClass={cs.CL},
      url={https://arxiv.org/abs/2607.02770}, 
}

@inproceedings{kwon2023efficient,
  title={Efficient Memory Management for Large Language Model Serving with PagedAttention},
  author={Woosuk Kwon and Zhuohan Li and Siyuan Zhuang and Ying Sheng and Lianmin Zheng and Cody Hao Yu and Joseph E. Gonzalez and Hao Zhang and Ion Stoica},
  booktitle={Proceedings of the ACM SIGOPS 29th Symposium on Operating Systems Principles},
  year={2023}
}
